\documentclass[11pt,a4paper,fleqn]{article}

\usepackage{graphicx}    
\usepackage{amsmath}   

\usepackage[T1]{fontenc}
\usepackage[latin1]{inputenc} 

\usepackage{xcolor}
\usepackage{amssymb}

\newcommand{\V}[1]{\mathbf{#1}}   
\newcommand{\M}[1]{\mathbf{#1}}   

\usepackage{fancyhdr}
\begin{document}

\title{Learning intuitive physics and one-shot imitation using state-action-prediction self-organizing maps}
\author{Martin Stetter$^1$ \and Elmar W. Lang$^2$}
\date{%
	\scriptsize
	$^1$Dept. of Bioengineering Sciences, Weihenstephan-Triesdorf University of Applied Sciences, Freising, Germany\\%
	$^2$CIML, Biophysics, University of Regensburg, Regensburg, Germany 
}
\maketitle

\begin{abstract}
Human learning and intelligence work differently from the supervised pattern recognition approach adopted in most deep learning architectures. Humans seem to learn rich representations by exploration and imitation, build causal models of the world, and use both to flexibly solve new tasks. We suggest a simple but effective unsupervised model which develops such characteristics. The agent learns to represent the dynamical physical properties of its environment by intrinsically motivated exploration, and performs inference on this representation to reach goals. For this, a set of self-organizing maps which represent state-action pairs is combined with a causal model for sequence prediction. The proposed system is evaluated in the \emph{cartpole} environment. After an initial phase of playful exploration, the agent can execute kinematic simulations of the environment's future, and use those for action planning. We demonstrate its performance on a set of several related, but different one-shot imitation tasks, which the agent flexibly solves in an active inference style.
\end{abstract}
\section{Introduction}

During the last decade, rapid progress in the field of deep learning has led to a number of remarkable achievements in many fields of artificial intelligence (AI) \cite{goodfellow2016}. However, human learning and intelligence seem to work radically differently from the supervised pattern recognition approach adopted in most deep learning architectures. Among many other things, humans, for example playing infants, are able to learn from exploration and imitation, learn from much fewer examples and create richer representations \cite{lake2017}. They can flexibly reason over these representations and creatively elicit novel state configurations never seen before. 

Here we suggest a simple neural network architecture which learns to represent the dynamic physical characteristics of its environment in an unsupervised, exploratory way. By inference on the basis of this representation the system can plan actions to reach externally given or intrinsically generated goals. In the following, we summarize related work and outline the proposed model.


Impressive performance of deep learning approaches has been demonstrated in some classical supervised tasks such as object  \cite{krizhevsky2012} or speech recognition \cite{graves2004, graves2013}, but also in unsupervised domains including representation learning and learning generative models \cite{goodfellow2014, kingma2013, razavi2019}. Recently, "deep reinforcement learning" \cite{mnih2015} has been demonstrated to achieve human or super-human level performance on playing Atari video games from raw pixel frames.   

Despite of all these achievements, studying deep learning might be less useful when it comes to understanding human-like intelligence with the goal to create artificial general intelligence \cite{lake2017}. Major issues raised include: 

\begin{itemize}
	\item  Deep learning approaches are in essence model-free and require massive amounts of labelled examples - orders of magnitude more than humans - in order to learn a complex task \cite{lake2017, rawlinson2019}.
	\item In most deep learning approaches, the original problem is being re-formulated in a clever way as a related, supervised task which can be tackled by deep multilayer perceptrons. Hence the intelligent, creative part is done by the human designer, not by the algorithm.
	\item Deep learning approaches work on associations and cannot grasp cause and effect \cite{pearl2018}.
\end{itemize}

The first issue has been addressed in many ways with approaches summarized as "few-shot learning" (for a recent review see \cite{wang_y2019}). Generally, a model is trained on a large body of related tasks to learn an inductive bias or prior, which is then exploited to solve the task at hand based on only one or few examples. Impressive recent achievements include one-shot imitation learning in robots \cite{duan2017} and meta-reinforcement learning \cite{wang_j2016}, the latter showing close relationships to biological reinforcement learning \cite{wang_j2018}.   

Yet, the systems trained are extremely complex in terms of the amount of parameters, and consequently the amount of data required, when summing over the different tasks, is still very high. 

Addressing the second and third issues seems to require a qualitative change in the paradigm of an artificial intelligent system, compared to deep learning. 

Recently, Lake, Ullman, Tenenbaum and Gershman \cite{lake2017} formulated cornerstones believed to be crucial ingredients of human-like learning and cognition, which is suggested to be more model-building like than pattern-recognition like: (i) "building causal models" of the world, (ii) "ground models on intuitive theories of physics and psychology", and (iii) "harness compositionality and learning-to-learn".  The authors state that in the approach of learning as model-building:
\begin{quote}
	"Cognition is about using these models to understand the world, to explain what we see, to imagine what could have happened and didn't, or what could be true and isn't, and then planning actions to make it so." (\cite{lake2017}, p.2)  
\end{quote}
Lake et al. \cite{lake2015} have developed "probabilistic program induction" as a model for few-shot visual concept learning, which focusses on compositionality and learning-to-learn.  

In our approach, which we outline in the next but one subsection, we address the first issue by deliberately designing a very simple i.e. strongly constrained yet biologically plausible model. Despite being simple, the model allows to address the second and third issue by being able to actively explore the causal structure of its environment, to learn intuitive physics and to plan actions in order to achieve goals.   


In this work we suggest a very simple neural architecture, 
which learns in completely unsupervised fashion and incorporates several of the mentioned principles: it learns a model of the dynamics of its environment by playful exploration ("intuitive physics"), can play virtual, predicted episodes ("what could be true and isn't") and can plan action sequences to bring the environment closer to a target state that has been given extrinsically by a one-shot demonstration ("planning actions to make it so"). Conceptually the same mechanism could be utilised, if the agent could generate intrinsic goals.

\begin{figure}[!h]
	\begin{center}
		\includegraphics[width=7cm]{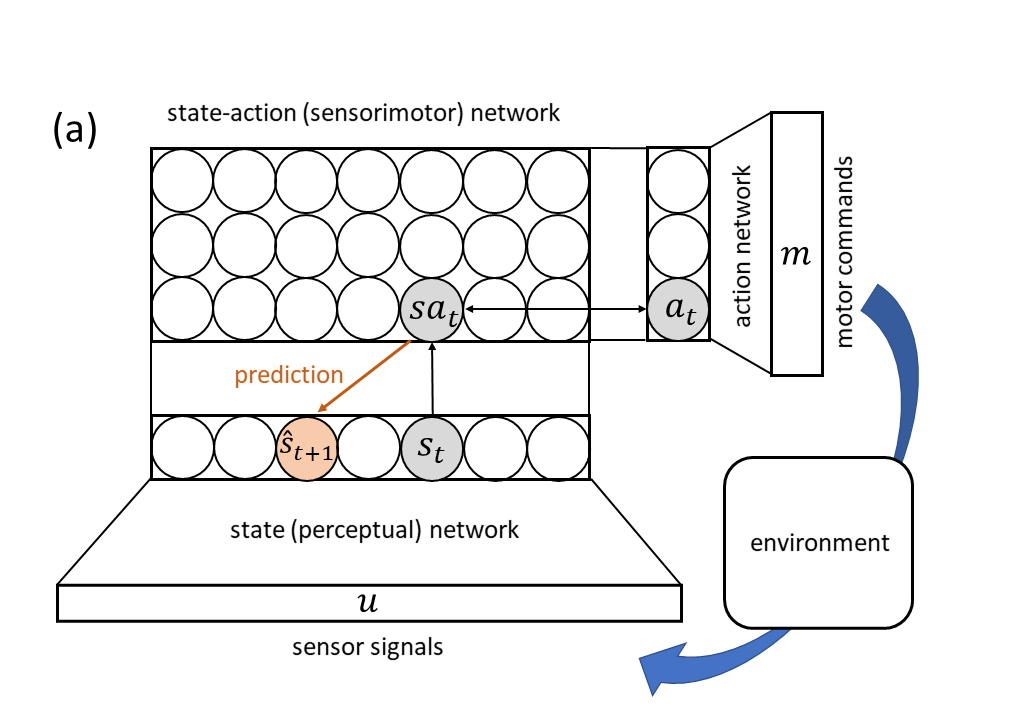}
		\includegraphics[width=7cm]{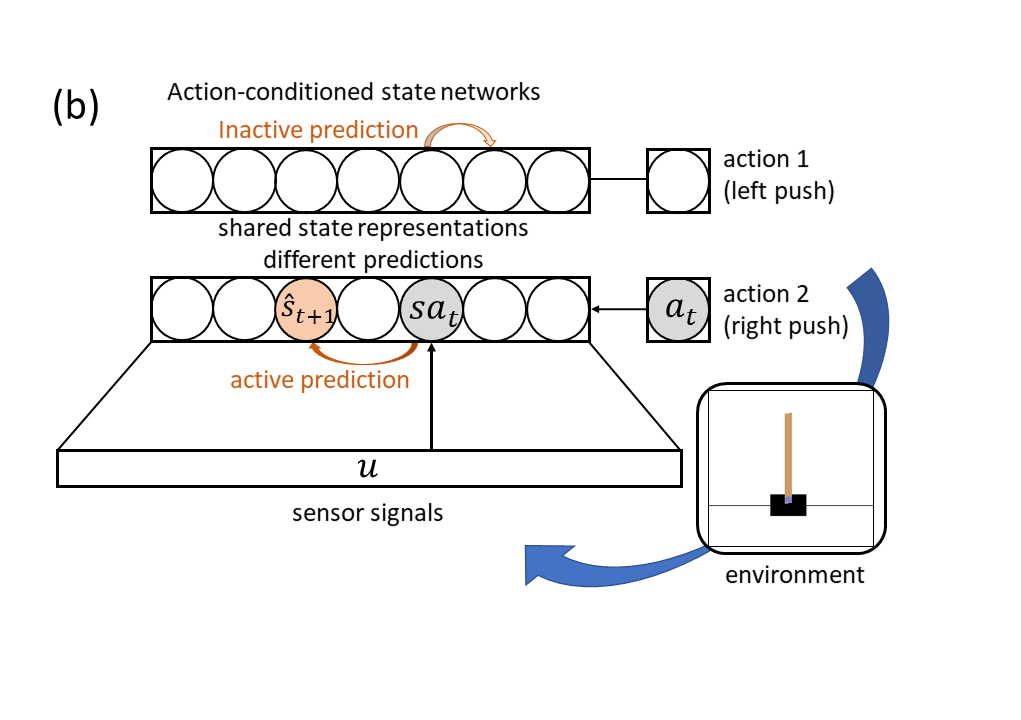}
	\end{center}
	
	\caption{ {\bf(a)} Proposed network architecture. A sensorimotor self-organizing map learns to represent state-action combinations, each represented by a state and action SOM, respectively. An activated state-action unit learns to predict the most likely next state, $\hat s_{t+1}$ (brown), conditioned on the current state and action, $s_t, a_t$ it represents. {\bf(b)} Reduced architecture actually implemented for the demonstrations in the result section (for details see text). 1D state and action representations are drawn for simplicity. }
	\label{fig:principle}
\end{figure}

The proposed system consists of sparse unsupervised networks, which in this work are implemented as Kohonen-type self-organizing maps (SOM) \cite{kohonen1982}: a perceptual or \emph{state} network, which learns a representation of the environment's states, an \emph{action} module which represents possible motor commands, and a sensorimotor integrating \emph{state-action} network which learns and represents associations between both (Fig \ref{fig:principle}a). During "playful exploration", the agent executes (e.g. randomly sampled) actions and observes resulting state changes. Observing means that each active state-action unit learns to predict the environment's next state given the currently active state-action pair it represents. For this, a transition model is learned by updating state transition matrices on top of the SOMs. This mechanism discovers the effects caused by the agent's own actions. For convenience, we refer to this architecture as "state-action-prediction self-organizing maps" (SapSom).  When proposing this architecture, the authors are well aware that mere temporal order does not necessarily reflect a true cause-effect relationship: the rooster crows before sunrise, but it does not cause sunrise. Nevertheless, the rooster's crow can be used to predict sunrise with a decent hit-rate.

In the proposed setup, playing virtual episodes or "kinematic mental simulation", for which evidence has been found in human reasoning \cite{khemlani2013}, corresponds to repeated prediction of state transitions starting from a virtual start state under a virtual action sequence. Here, the term "virtual" denotes intrinsically generated states,  which are represented by active units without sensory stimulation and without action to be executed. This process of playing virtual episodes serves to transform a virtual action sequence into a predicted sequence of states. 

A goal is defined as a target state or group of states in latent state space. Action planning corresponds to searching in action sequence space in the style of active inference,  such that predicted resulting states approximate as good as possible a desired region in latent space (i.e. reach that goal). 

The active inference paradigm adopted here \cite{friston2003, friston2005, friston2010} represents the action generation mechanism proposed in the predictive processing (PP) paradigm of brain modelling \cite{rao1998, hohwy2013, clark2016}. PP models view the cerebral cortex as a hierarchically organized prediction engine which constantly tries to explain i.e. predict, incoming sensory data as effect of hidden causes. Successful prediction is suggested to cause the subjective percept. Hence, sensor signals act as supervisory signals, and the brain's internal states (hypothesized causes) play the role of generative signals or inputs, which are adjusted such as to minimize prediction error.
Active inference refers to the process of acting on the environment such as to make the own prediction come true. 


The paper is organized as follows: An introductory section on related work is followed by the Model section, where we specify the model architecture including the representation and predicition learning procedures, and the mechanism of inference and action planning. 
In the Method and Results section thereafter we provide a proof of concept using the Open AI's gym cartpole environment. The section is subdivided in three parts: First, SapSom is trained by playfully exploring the cartpole environment using random action sequences. It is shown that the agent correctly learns to represent the phase space structure of the cartpole system, referred to as "intuitive physics". In the second part, the ability to perform kinematic simulation is examined by comparing real and simulated dynamics of the environment under three pre-specified deterministic action sequences as well as under many random action sequences.  A quantitative error measure is provided which demonstrates that predicted state sequences closely resemble the actual temporal state sequences provided by the environment. In the third part of the Method and Results section we analyze the ability of the agent to reach various goals, which are extrinsically given as successful example state sequences. Three classes of goals are considered - the balancing task, the controlled-tilt task, and the tilted-balancing task - and individual runs for visual inspection as well as a quantitative performance assessment are given. It is shown that the agent flexibly and most of the time successfully solves many instances of these tasks after observing only one demonstration. We conclude the paper by discussing how our model might be extended to yield more powerful architectures, and relating our model to biological findings. A preprint of this work has been made available under \cite{stetter2021}.

\section{Related work}

In contrast to current trends in deep learning, human-like learning systems need to build causal models of their environments (causality), acquire an intuitive understanding of the underlying physics (intuitive physics) and develop learning-to-learn skills and combining constituents (compositionality) to generalize knowledge to new tasks and events \cite{lake2017}. As such, learning intuitive physics has become a major target of recent research activities in cognitive sciences and artificial intelligence. Humans seem to have a basic understanding of objects and their physical inter-actions from infancy \cite{spelke2007}, which becomes refined during development by self-curated experiments with their environments. This curiosity-driven learning of intuitive physics \cite{schmidhuber2010} helps humans to mentally run physics simulations \cite{battaglia2013} in order to plan and predict future actions and events. Furthermore, a task-to-task transfer and generalization to unseen events remains a big challenge to any artificial intelligence system.

The SapSom model presented below emphasizes simplicity in the sense of strongly constrained and therefore easy-to-train architecture. A set of SOMs is equipped with a simple yet powerful additive, Hebb-trained prediction system. In spirit it shows close relationship to model-predictive control. There are similar  approaches which address intuitive physics learning on the basis of self-organizing maps. For example Toussaint \cite{toussaint2006} designed  a sensorimotor SOM, in which lateral connections of a sensory map are modified by motor activations in a multiplicative, modulatory way. Morasso et al. \cite{morasso1997} address the targetting movement problem in robotics by setting up one joint SOM for sensory and motor spaces. In contrast to these approaches, we use several SOMs with action-conditioned transition models, which in the consequence can become much simpler to train than in the mentioned approaches. Apart from this class of approaches with limited complexity, a large number of intuitive physics learning and planning systems have been proposed, which show relationships to deep learning. 

A common approach to learn intuitive physics is via handcrafted simulation engines \cite{battaglia2013} or via physics engines made adaptable through deep neural networks \cite{wu2015, battaglia2016}. More generally, dynamics of object interactions have been learned either jointly with perception \cite{watters2017} or through decomposition of visual scenes into structured representations of objects and their dynamic states \cite{chang2017, vansteenkiste2018}.  Alternatively, perception modules have been further developed to be able to either imagine future states of objects \cite{fragkiadaki2017} or to predict them through a GAN - based approach or an encoder-decoder network \cite{riochet2018}. With a similar goal, a graph-based perception-prediction-network (PPN) has been proposed in \cite{zheng2018} to learn without human intervention latent object properties from their interactions. During a gradient-based training with samples of object dynamics, the perception module generates representations of object properties, while a prediction module uses the latter to simulate system dynamics. The PPN seamlessly generalizes to unseen scenarios, and its learned representations can be translated into human - interpretable properties. As discussed in \cite{battaglia2018}, structured representations through graph networks provide a strong relational inductive bias to support reasoning and combinatorial generalization. The perception-prediction concept has also been harnessed by Wu et al. \cite{wu2017} to learn intuitive physics. There a perception module first learns a representation of the physical environment, which is then used by physics and graphics engines to learn interpreting and reconstructing the visual stimulus sequences. Later the generative models are used for reasoning and prediction. Inverting a physics engine is achieved by a convolutional inversion network.

As sensor signals are often fraught with uncertainty, models of the outside world need to integrate sensory inputs with internal prior beliefs, in accordance with Bayesian inference. Hence probabilistic approaches are able to deal with uncertainty and, combined with system identification techniques like particle filters achieve robust long-term predictions of object dynamics while simultaneously learning the underlying intuitive physics \cite{ehrhardt2017, smith2019}.

Several recent approaches refer to learning physics from observations alone. Applying exploratory learning of causal relationships was the intention of Kansky et al. \cite{kansky2017} when designing an object-oriented generative physics simulator. It is based on a complex structured network architecture and can learn the dynamics of an environment only from observations. By reasoning backward through causes this Schema Network can transfer intuitively learnt physics to unseen situations. An unsupervised learning of intuitive physics purely from visual observations was  reported in \cite{ehrhardt2019}. Unsupervised predictors of physical states were constructed by tracking salient objects in dynamic stimulus sequences based on causality and equivariance, and using the learnt physical states to train visual predictors, which successfully could incorporate the underlying environment. With a similar goal, Ehrhardt et al. \cite{ehrhardt2019b} implemented an unsupervised meta-learning formulation to predict the dynamics of moving objects in a complex environment and to generalize to a varying number of moving objects. The network architecture comprises sequential auto-encoders as well as an U-net encoder-decoder to extract the relevant information. Finally an image sequence is predicted employing a stack fully convolutional network, while the parameters of the system are adapted by optimizing a proper loss function. Meta-learning was also used to train neural networks augmented with  an auto-encoder-based meta-recognition model producing model code, which a subsequent meta-generative model uses to construct parameters for task-specific models. This meta-learning auto-encoder (MeLA) framework is able to build models for previously unseen tasks, which closely match the true underlying models \cite{wu_t2018}. A simulator-augmented interaction networks (SAIN) model was proposed in \cite{ajay2019}, which combines object-based networks, which learn residuals, with an object-based learnable physics engine to search for actions in control problems.

Rather than relying on physics engines, Nguyen et al. \cite{nguyen2020} design a surprise and explain (SnE) framework, which rests upon the basic premise that object dynamics is mostly linear, and any non-linear dynamics presents a surprise for which an explanation cannot be inferred easily. The anomaly detection framework comprises three modules: perception, dynamics and explanation. An interpretable intuitive physics model has also been designed by Ye et al. \cite{ye2018}. The bottleneck layers of the deep network architecture contain specific dimensions, which correspond to different physical properties. The model is trained with sequences of colliding objects and generalizes well to scenarios with different underlying physical properties.

\section{Model}

\label{sec:model}
In this work we actually implement a reduced version of the model outlined in Fig \ref{fig:principle}a. The resulting simplified architecture is shown in Fig \ref{fig:principle}b. Simplifications are: (i) no action map is explicitely learned, which would maintain codebook vectors of motor commands. This restricts the implementation to discrete finite action spaces, which is, however, sufficient for the cases studied here. (ii) instead of a full sensorimotor map, a set of $K$ action-conditioned state-only maps is trained, where $K$ is the number of actions.  All maps share the same perceptual (state) representation, but on top of each map an individual, action-conditioned state transition matrix is learned. This prohibits
dimensional control of the sensorimotor space, but makes sure that all state action pairs are readily represented.  

\subsection{Representation learning}

We wish to learn a sparse representation of the input space, because it is believed that this will facilitate the learning of unimodal sharply peaked state transition distributions. As discussed above, many powerful techniques for representation learning and even several possibilites for obtaining sparse representations are around (e.g. \cite{foeldiak1990}). Here we use self-organizing maps (SOMs) \cite{kohonen1982}, because in addition to sparsity they maintain topographic order in map space, which may be useful in terms of predictive processing. 

SOMs have been successfully used both in bio-inspired hierarchical representation learning \cite{miller2006} and reinforcement learning \cite{rawlinson2012}. To briefly summarize, in a SOM, neurons are geometrically arranged in a regular grid (here 2D rectangular). Each unit at map location $ \V{s} \in \mathbb{R}^2$ maintains a codebook vector $\V{w}(\V{s})$ with the same dimension as the input space. On presentation of an input vector or "environmental state" $\V{u}$, the winning unit $\V{s}^*$ is found, defined by its codebook vector being closest to the input 

\begin{equation}
	\V{s}^*(\V{u}) = \text{argmin}_{\V{s}} ||\V{u} - \V{w}(\V{s})||,
	\label{eq:winner}
\end{equation} 

where $|| \cdot ||$ denotes euclidean norm, and the notation on the left hand side of eq. (\ref{eq:winner}) explicates the dependence of the winner on $\V{u}$. The winning unit and its neighbours in map space learn according to 

\begin{equation}
	\Delta \V{w}(\V{s}) = \eta \; h_{\V{s}^*(\V{u})}(\V{s}) \; (\V{u} - \V{w}(\V{s})), \qquad  h_{\V{s}^*(\V{u})}(\V{s}) = \exp \left(-\frac{||\V{s}^*(\V{u}) - \V{s}||^2}{2\sigma^2}\right)
	\label{eq:learning}
\end{equation}

where $h$ can be interpreted as a localized neural activation pattern centered around the winning unit and $\sigma$ is defined in eq. (\ref{eq:sigma}) below. The strong localization of activation entails that only a small fraction of units is active at every time step, resulting in a sparse code.
The learning rule eq. (\ref{eq:learning}) brings codebook vectors of map neighbours both closer to regions of high data density and to each other. 

\subsubsection{SOMs and predictive processing}
There is an interpretation of SOMs in terms of predictive processing: when considering the codebook vector of an active unit $\V{s}$  as prediction of the input, $||\V{u} - \V{w}(\V{s})||$ is just the prediction error in input space (the term "prediction" being used in the sense of "predicting the presence"), often referred to as quantization error. Finding $\V{s}^*(\V{u})$ according to eq. (\ref{eq:winner}) thus minimizes the input prediction error by inference, and the learning step eq. (\ref{eq:learning}) further minimizes it by learning. 

Similarly to \cite{miller2006}, we define a data-driven variable width $\sigma$ of the activation pattern, namely

\begin{equation}
	\sigma = \sigma_0 \dfrac{\min_{\V{s}}||\V{u}-\V{w}(\V{s})||}{\text{mean}_\V{s}||\V{u}-\V{w}(\V{s})||},
	\label{eq:sigma}
\end{equation} 

where $\sigma_0$ has been empirically set to 25 percent of the map diameter throughout this work. 
By this, for inputs that are well-represented with small prediction error, only a small neighbourhood learns, whereas a large prediction error (maybe due to nonstationary statistics of the input) causes a large portion of the map to rearrange to better represent this novel input. 

The normalized activation pattern in map space can then be interpreted as recognition density, where 

\begin{equation}
	p(\V{s}|\V{u}) =  \frac{h_{\V{s}^*(\V{u})}(\V{s})}{\sum_{\V{s'}} h_{\V{s}^*(\V{u})}(\V{s'})}.
	\label{eq:prob}
\end{equation}

According to eq. (\ref{eq:sigma}), the recognition density is sharply peaked for small prediction errors and more distributed for larger representational uncertainty. In SOMs, the generative distribution, which denotes the likelihood of inputs given a map state, simply becomes $p(\V{u}|\V{s}) = \delta(\V{u} - \V{w}(\V{s}))$, $\delta$ denoting Kronecker's delta. For a comprehensive treatment of generative and recognition models, see e.g. \cite{friston2003, friston2005, friston2010}.

Finally, due to topographic order, prediction error in map space can be defined in geometrical terms, simply as geometric distance between most likely true and predicted states, respectively. 

\subsection{Prediction learning}

During learning, the system will exploit the possibility to act on the environment and to directly observe the consequences of these own actions on environmental state changes. Hence, learning is situated at the \emph{intervention} level of Pearl's Causal Hierarchy and can be formulated in the framework of \emph{Do}-calculus \cite{pearl2000}. Due to its similarity to how infants explore their environment by testing the effects of their actions, we metaphorically refer to this style of learning as "playful exploration".

During playful exploration, the model learns to approximate the action-conditioned Markov transition distribution $p(\V{s'} | \V{s}, do(a))$, where $\V{s'},\V{s}$ run over all state units and $a$ runs over all actions. 
The distribution denotes the probability of finding map state $\V{s'}$ activated one time step after in state $\V{s}$ action $a$ has been executed.
For this, each action-conditioned state network, labelled by $a$, updates an individual state transition matrix $\M{T}_a$ with components $T_a(\V{s'},\V{s})$ (for matrix operations, indices $\V{s'}, \V{s}$ being appropriately re-ordered as scalars), whenever $a$ is executed. Let $\V{p}_t$ be the column vector of probabilities $p(\V{s}|\V{u}_t)$, eq. (\ref{eq:prob}), assigned to map states at time $t$ in response to input $\V{u}_t$. On execution of action $a_t$, the environment changes to state $\V{u}_{t+1}$ leading to a new distribution $\V{p}_{t+1}$. The state distribution \emph{predicted} by network $a_t$, in contrast,  is given by $\V{\hat p}_{t+1} = \M{T}_{a_t} \cdot \V{p}_t$. We adopt a simple least squares scheme and mimimize $||\V{p}_{t+1} - \M{T}_{a_t} \cdot \V{p}_t||^2$ with respect to $\M{T}_{a_t}$. Gradient descent leads to the learning rule 

\begin{equation}
	\Delta \M{T}_{a_t} = \gamma\; (\V{p}_{t+1} - \V{\hat p}_{t+1}) \;\V{p}_t^T
	\label{eq:predlearn}
\end{equation}     

where $\gamma$ is a learning step size variable and the superscript $(\cdot)^T$ denotes the transpose of a vector or matrix. We prefer least squares over minimization of the Kullback Leibler divergence between both distributions, which is often pursued, because the latter is usually tractable only under severe simplifications, which often lead to treatment of modes only.

\subsection{Inference}

After exploring the environment to a sufficient extent, inference can be done on the so-far learnt representation using the state transition matrices. For example, given a certain start map state $\V{s}_0$, the system can generate virtual action sequences by activating action nodes due to some schedule, without actually executing the corresponding motor commands, and predict the sequence of states that would result from executing that sequence. Correlates of this in human cognition might be kinematic mental simulation with the goal to plan actions. Moreover, the start state might be virtually generated as well, instead of perceptually caused, giving the possibility to elaborate on virtual scenarios never seen before, which might be considered a kind of artificial creativity.
We do not formulate an explicit neuronal model of how state and action representations might be spontaneously generated, but there are mechanisms and models of how this might occur  spontaneously or in response to stimulation \cite{stetter2006}. In the present context, sequence prediction is referred to as "playing virtual episodes", as this procedures accepts a start state and an action sequence and returns a sequence of predicted environmental states. 

\subsubsection{Sequence prediction}

One step prediction given environmental state $\V{u}_t$ and action $a_t$ was done as "prediction by mode":  Given the currently winning unit, $\V{s}^*(\V{u}_t)$, the most likely next map state is calculated as $\V{\hat s}_{t+1} = \text{argmax}_{\V{s'}} \left( T_{a_t}(\V{s'}, \V{s}^*(\V{u}_t) \right)$, resulting in $\V{\hat u}_{t+1} = \V{w}(\V{\hat s}_{t+1})$. In order to avoid deadlocks, prediction of the currently winning state is suppressed. A sequence of environmental states given a start state $\V{u}_0$ and an action sequence is predicted by consecutively applying one-step predictions on the basis of estimated environmental states i.e.  first state $\V{\hat u}_1$ is predicted from the start state $\V{u}_0$ and $a_0$, and each next environmental state $\V{\hat u}_{t+1}$ on the basis of $\V{\hat u}_t$ and $a_t$. 

\subsubsection{Goals and action planning}

Sequence predictions can be used to plan action sequences in order to reach a goal. This requires the definition of what a goal is in the present context. We suggest to define a goal as a target state or a subset of target states in map space. Target states might be provided by stimulation e.g. by demonstrating a target environmental state to the system, or might be intrinsically generated as described in the previous paragraph. These target states are then imprinted or memorized, while the system tries to reach and maintain them by executing a suitable sequence of actions. A biological correlate of target state memorization might be persistent non-distractible neural activity found in prefrontal cortex \cite{fuster1971}. The described procedure is closely related to the following concepts: (i) one shot imitation: imprinting the target state corresponds to the single demonstration of the goal, reaching the goal is then done by inference over the learned intuitive physical model. (ii) active inference: a system predicts a target state and minimizes prediction error by driving the environment towards the predicted (i.e. desired) state.   

A large body of reinforcement learning literature exists on how to find a policy $p(a) = \pi(\V{s})$, which specifies how actions should be planned in order to maximize reward. Here we suggest an action planning strategy which does not rely on external reward signals but operates entirely on the distances between target states and the current state. 
Actually, the drive to try and reach an imprinted goal representation by active inference must in some respect be generated by an intrinsic reward mechanism, which is, however, not explicitly modelled here. Possible distance measures include 
(here euclidean) distance either between environmental states, $||\V{u}^{\text{target}} - \V{u}_t||$, which is the input prediction error in active inference terminology, or - because of topographic order - between map states, $||\V{s}^{\text{target}} - \V{s}_t||$, or both. We found action plans on the basis of environmental state distances to work better than map distance for the tasks considered here, hence the results in the Method and Results section are generated using this distance measure.

Many distance-based action planning schemes can be imagined, here we use simple $\tau$ step greedy forward search for action planning: On the basis of the true present state $\V{u}_0$, find the action sequence of length $\tau$, the execution of which minimizes the distance between the predicted state resulting from that action sequence and the target state. 

\section{Method and Results}
\label{sec:results}
The system was implemented in Python, Version 3.7.6 64 bit, using PyTorch's tensor library  version 1.4.0, \texttt{https://pytorch.org/}. All experiments have been carried out on a desktop PC equipped with an Intel(R) Xeon(R) E-2186G CPU (6 cores, 3.80GHz) and an NVIDIA Quadro P1000 graphics board. SapSom was tested on the Open AI gym cartpole environment (for a screenshot of the rendered cartpole see Fig \ref{fig:principle}b, inset), \texttt{https://gym.openai.com/envs/CartPole-v0/}. Hence, no external data sets were used, the only data were the variables returned by the gym environment in response to the agent's actions. The SapSom source code has been opensourced under \texttt{https://github.com/martinstetter/sapsom}. 

For the experiments shown, a $16 \times 16$ SOM was trained and analyzed as specified below. The environment accepts two actions, push the cart to the left ($a_1$) or to the right ($a_2$) with a fixed force, and emits four sensory signals, namely the cart location $x$, the pole's angle with the vertical $\theta$ and their temporal changes i.e. $\V{u} = (x, \dot x, \theta, \dot \theta)$. The system was originally designed as a testbed for reinforcement learning systems with the goal to keep the pole vertical by balancing, therefore the enviroment also returns a reward for each step and triggers a "done" signal, as soon as the pole hits $\pm 15$ degrees, the cart hits the screen border, or 200 steps of balancing are successfully executed. Throughout this work, the reward signal was ignored, because SapSom operates in a completely unsupervised way. A sequence of steps between cartpole initialization and trigger of the done signal is referred to as an episode.

The system was trained as follows: in order to assure correct unfolding of the map, the SOM representing the input space was pretrained over $1000$ episodes under random action sequences using a standard learning scheme for self-organizing maps with exponential decays for $\sigma$ and $\eta$ between start and end values $(8, 0.1)$ and $(0.3, 0.01)$, respectively. 
Subsequently, both representation and prediction parts were trained simultaneously over $3000$ episodes  with $\sigma_0 = 4, \eta_0 = \gamma = 0.05$ as follows: for a given state first the SOMs were updated under adaptive neighbourhood, then an action was selected and executed, the resulting next state was observed, and the corresponding transition matrix was updated as specified in the model section. Randomly selected actions were used during "playful exploration".

\subsection{Intuitive physics}
Here we tested whether SapSom could learn a representation of the environment's Newtonian dynamics, metaphorically referred to as "intuitive physics" \cite{lake2017}. In technical terms, we tested whether, after training,  the system could approximate the real phase portrait of the environment by its own predicted phase portrait. Results are shown for the $\theta - \dot \theta$ phase plane, because the pole's behaviour rather than the cart's behaviour is usually considered in the cartpole environment.

After training, the real and predicted dynamics of the environment were analyzed by playing real episodes and determining the real and predicted directions of motion from one step to the next. For a given phase point $\V{u}_t$ and next action $a_t$, the real direction of motion was determined by executing $a_t$ on the environment and calculating $\V{u}_{t+1}-\V{u}_t$. The predicted direction of motion was calculated by applying $\V{u}_t$ and determining $\V{p}_t$, then predicting the next state $\V{\hat s}_{t+1}$ and corresponding predicted input $\V{\hat u}_{t+1}$, and finally computing $\V{\hat u}_{t+1} - \V{u}_t$. 

\begin{figure}[!h]
	
	\begin{center}
		\includegraphics[width=7.6cm]{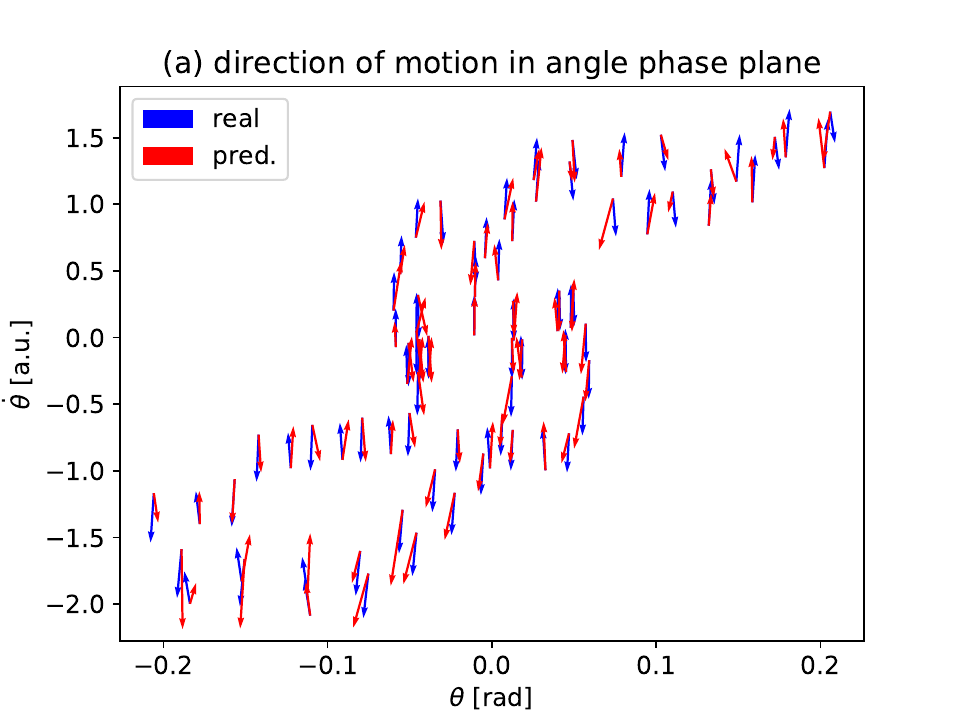}
		\includegraphics[width=7.6cm]{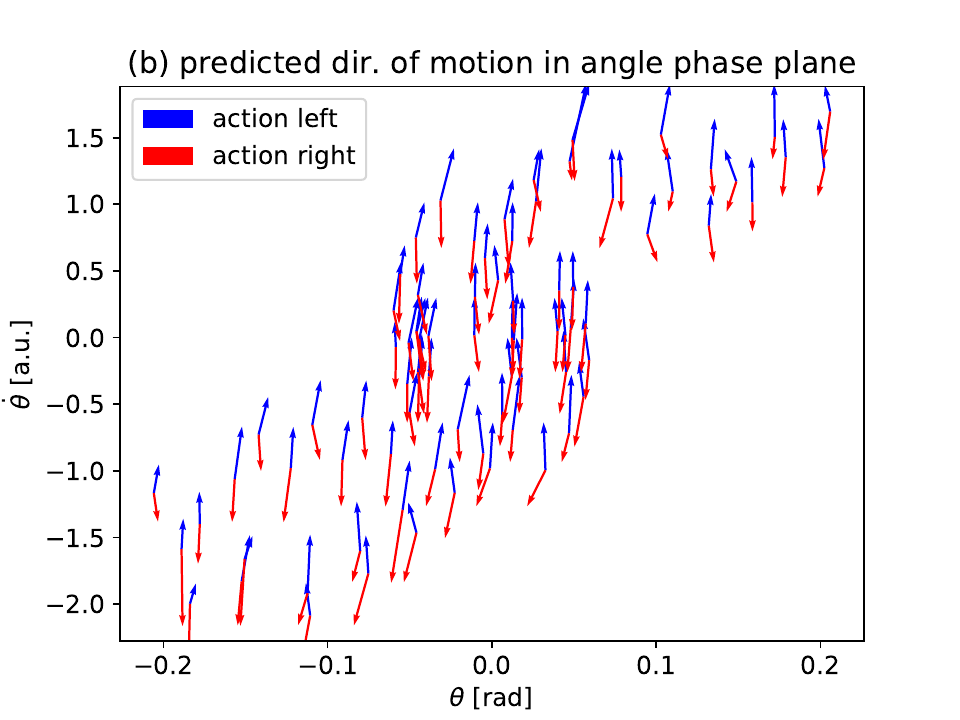}
	\end{center}
	\caption{Directions of motion in the $\theta - \dot \theta$ phase plane (arrows) for five complete random episodes. Arrows are located at the states $(\theta, \dot \theta)$ to which they apply.  {\bf(a)} Blue: Real directions of motion in the next step as provided by the environment. Red: Directions of motion predicted by the network when provided with the same state and action. Predictions approximate real movements very well. {\bf(b)} Prediction of motions in phase space under virtual left push (blue) and virtual right push (red), respectively (for discussion see text). }
	\label{fig:quiver}
\end{figure}

The real (blue) and predicted (red) directions of motion in the angle phase plane are shown in Fig \ref{fig:quiver}a for five complete random episodes. Note that this phase portrait is not uniquely defined, because at each point directions are conditioned on $a, x$ and $\dot x$. Real and predicted directions of motion agree very well with each other. However, there are small deviations, although in principle the cartpole physics is deterministic and should in principle be learnable to arbitrary accuracy. The existence of small deviations is due to the state representation's quantization error: similar, but different environmental state trajectories will be mapped to the same map unit, but will have slightly different time evolutions. These differences cannot be resolved by the system. Where quantization errors become large (e.g. for novel states), prediction errors can become large as well.    

Fig \ref{fig:quiver}b displays SapSom's predicted directions of motion when planning to execute a left push (blue) or a right push (red), respectively, for the same five episodes. The configuration reflects correct "comprehension" of the situation: a left push generally accelerates the pole to the right i.e. angular velocity increases, which is correctly mirrored by the blue arrows pointing upwards towards increasing $\dot \theta$. Under opposite sign, the same is true for right push.

We conclude, that SapSom can learn a reasonably accurate representation of the cartpole dynamics only from interventional exploration. Because this is achieved in a completely unsupervised way and without making explicit use of the equations of motion, this can be understood as a way of capturing an intuition about the physics of the environment.  

In the next two subsections, we present results about inference on this model. In order to separate slow learning and inference effects, learning was switched off in the following by setting $\eta_0 = \gamma = 0$.

\subsection{Playing virtual episodes}

Next we examined whether the trained system could virtually play episodes on the basis of its intuitive physics i.e. whether given an action sequence and a start state, the corresponding future state sequence could be predicted reasonably well.

\begin{figure}[h!]
	\begin{center}
		\includegraphics[width=9cm]{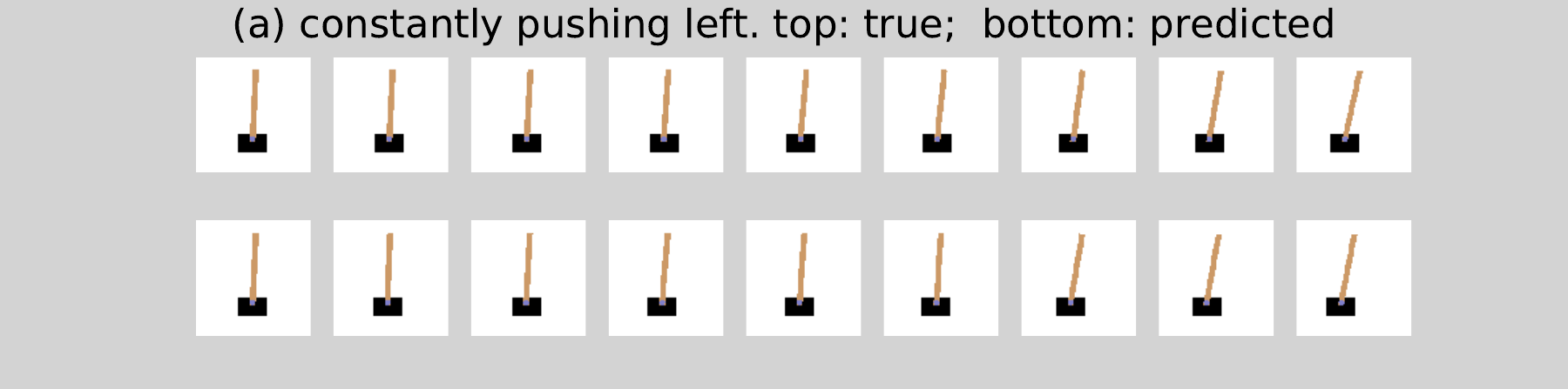}\\[-2mm]
		\includegraphics[width=11cm]{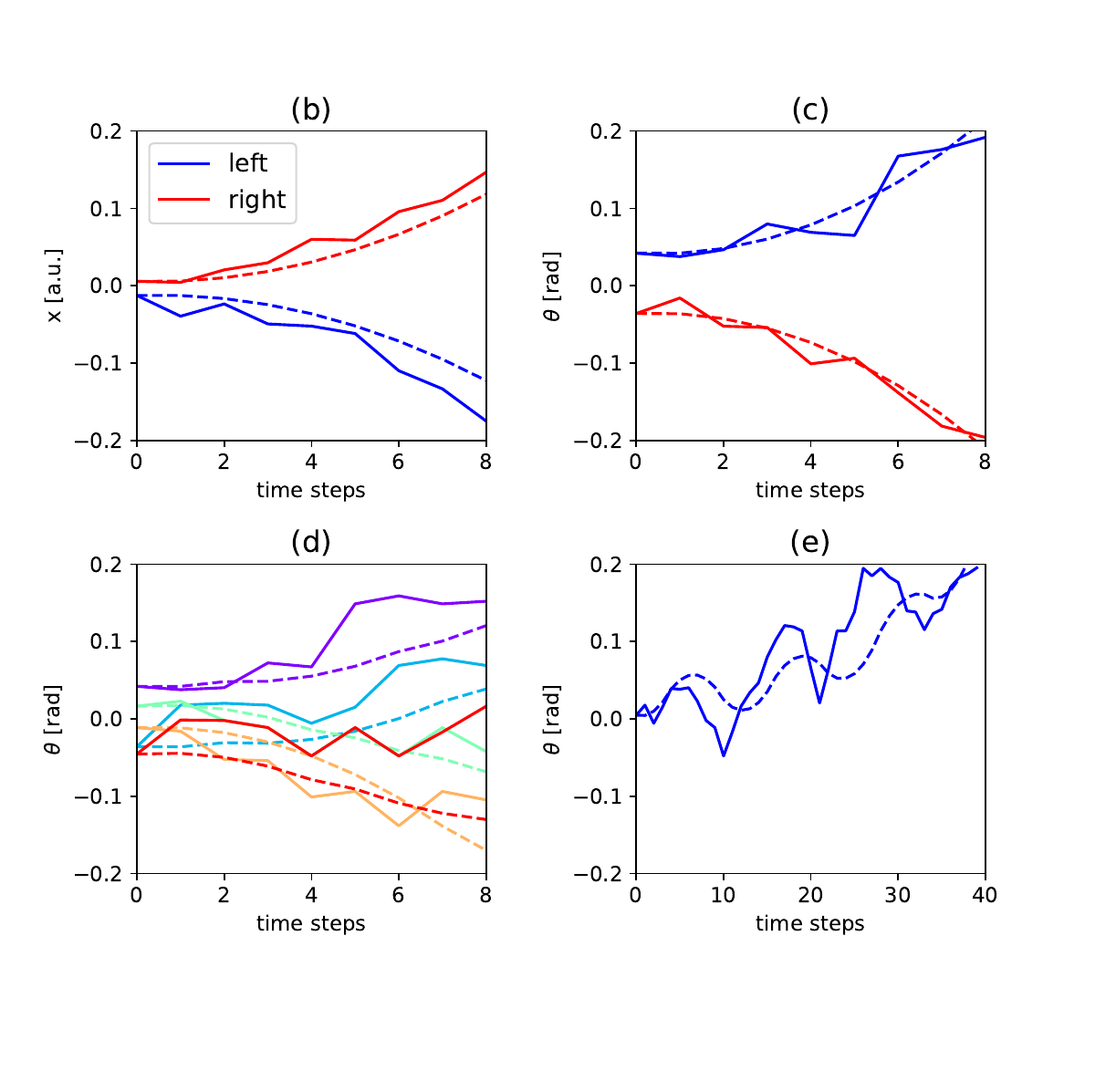}
	\end{center}
	\vspace*{-1cm}
	\caption{{\bf(a)} Screen shots of cartpole with identical start state followed by eight left pushes. top: real time evolution. bottom: 8-step prediction of time evolution. {\bf(b-e)} Real (dashed) and predicted (solid) time evolutions of $\theta$ and $x$ starting from identical initial states. {\bf (b)} Time evolution of $x$, and {\bf (c)} time evolution of $\theta$ for the same simulation run. Blue: 8 left pushes; red: 8 right pushes {\bf (d)} Time evolution of $\theta$ for 5 different rendom action sequences and cartpole initializations. Traces with same colors correspond to the same simulation run. {\bf(e)} Time evolution of $\theta$ for a longer sequence of length 39: oscillatory actions (3 left followed by alternating (6 x right) (6 x left) pushes). Major features of motion are correctly captured in all cases. 
	}
	\label{fig:imagined}
\end{figure}

Results for a number of different scenarios are summarized in Fig \ref{fig:imagined}. The top Fig \ref{fig:imagined}a illustrates by a number of screenshots, how the cartpole evolves under its true dynamics (top row) and under predicted dynamics for the same start state and action sequence (bottom row). Actions were eight left pushes. The bottom row images were generated by manually setting the cartpole's state to $\V{\hat u}_t$, rendering the environment, and then capturing the screen. From visual inspection one may conclude that both sequences agree very well. 

A slightly more quantitative analysis of prediction quality is given by plotting the time evolution of $x$ (in arbitrary units as provided by the gym environment) 
and $\theta$ under true (dashed) and predicted (solid) dynamics. 
Fig \ref{fig:imagined}b  and Fig \ref{fig:imagined}c show how $x$ and $\theta$ evolve over time, where blue traces correspond to eight left pushes and red traces to eight right pushes.
Fig. \ref{fig:imagined}d displays the time evolution of $\theta$ under five different random action sequences and initial conditions of the cartpole. Corresponding real and simulated dynamics are plotted in the same colour. Although the coincidence is not perfect, the general features of the resulting motions (shifting and tilting to the correct direction) are captured quite well. 
In order to quantify the prediction quality, simulated and true dynamics were generated repeatedly under 100 different random action sequences and cartpole initializations, and the RMSE between real and predicted angle was computed per time step of prediction horizon. The results are summarized in Table \ref{tab:rmse}. The values moderately increase from 0.028 rad to 0.0423 rad as the number of time steps into the future increases, corresponding to a relative error of 15 to 20 percent when comparing to the absolute range of $\theta$ values.

\begin{table}[h!]
	\caption{RMSE of predicted angle over time steps into the future}
	\begin{center}
		
		\begin{tabular}{|c||c|c|c|c|c|c|c|}
			\hline
			t & 1 & 2 & 3 & 4 & 5 & 6 & 7\\
			\hline
			RMSE & 0.0280 & 0.0293 & 0.0315 & 0.0337 & 0.0396 & 0.0414 & 0.0423\\
			\hline
		\end{tabular}
	\end{center}
	\label{tab:rmse}
\end{table}

Finally, in order to visualize the prediction performance on a longer and more complicated sequence, the comparison was run under an action sequence of length $39$, which elicits an oscillation (Fig \ref{fig:imagined}e). 
The gym environment initializes the cartpole's start state with small random numbers, resulting in a small positive initial value for $\theta$ in this case. As a consequence, in the true dynamics (dashed line) a slowly accelerating tilt to the right, in the direction of increasing $\theta$, under gravity  is superimposed with the faster oscillation evoked by the action sequence. 

The general behaviour (increasing $\theta$ under oscillation) is correctly predicted by the agent (solid line), even though the difference between the absolute values of real and predicted angles increases over time as indicated also in Table \ref{tab:rmse}. This increasing prediction error can be understood by keeping in mind that, besides the action sequence, only the start state is available to the system (i.e. no intermediate sync). 

It may be concluded that, for the environment considered, SapSom can perform qualitatively and semi-quantitatively correct multistep predictions of the environment's future under a given virtual action sequence (usually generated by the agent itself). This encourages us to test the system's action planning performance when performing a task. Since in the present system controling the environment in order to achieve a goal is an inferential rather than slow learning process, we test SapSom on a set of one-shot imitation tasks.


\subsection{One-shot imitation}

A task requires a goal to be formulated, either implicitely (by reward structure) or explicitely, by demonstrating one or a few success stories, i.e. examples where the goal has been reached. Here we adopt the latter approach, which has the advantage that no sometimes complex reward structure needs to be formulated. For SapSom, we define a goal as a target state or set of target states in its state representation, which it has to reach and maintain. This target state can be spontaneously generated by the system (intrinsic or curiosity-driven goal) or can be imprinted from outside (extrinsic goal).

\begin{figure}[h!]
	\begin{center}
		\includegraphics[width=13cm]{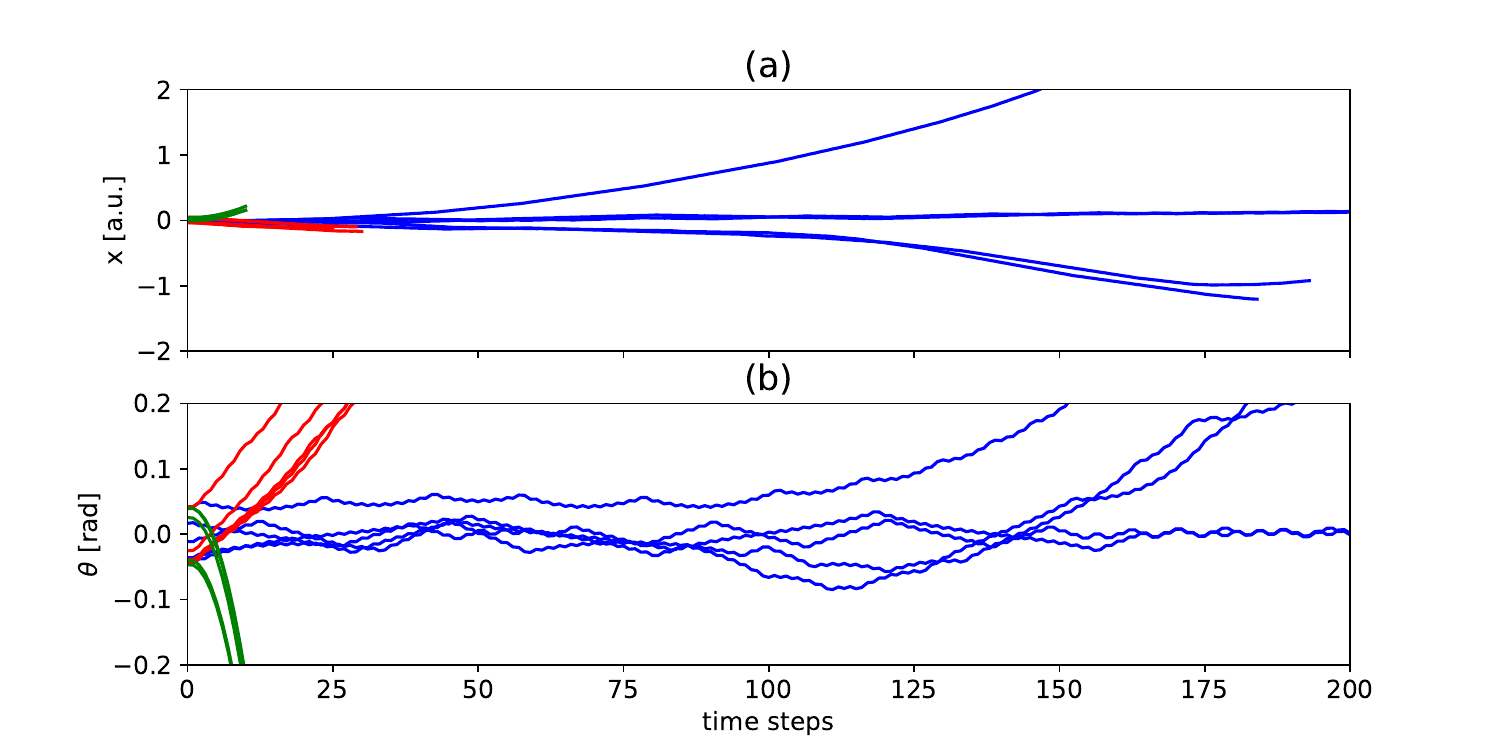}\\[2mm]
		\includegraphics[width=13cm]{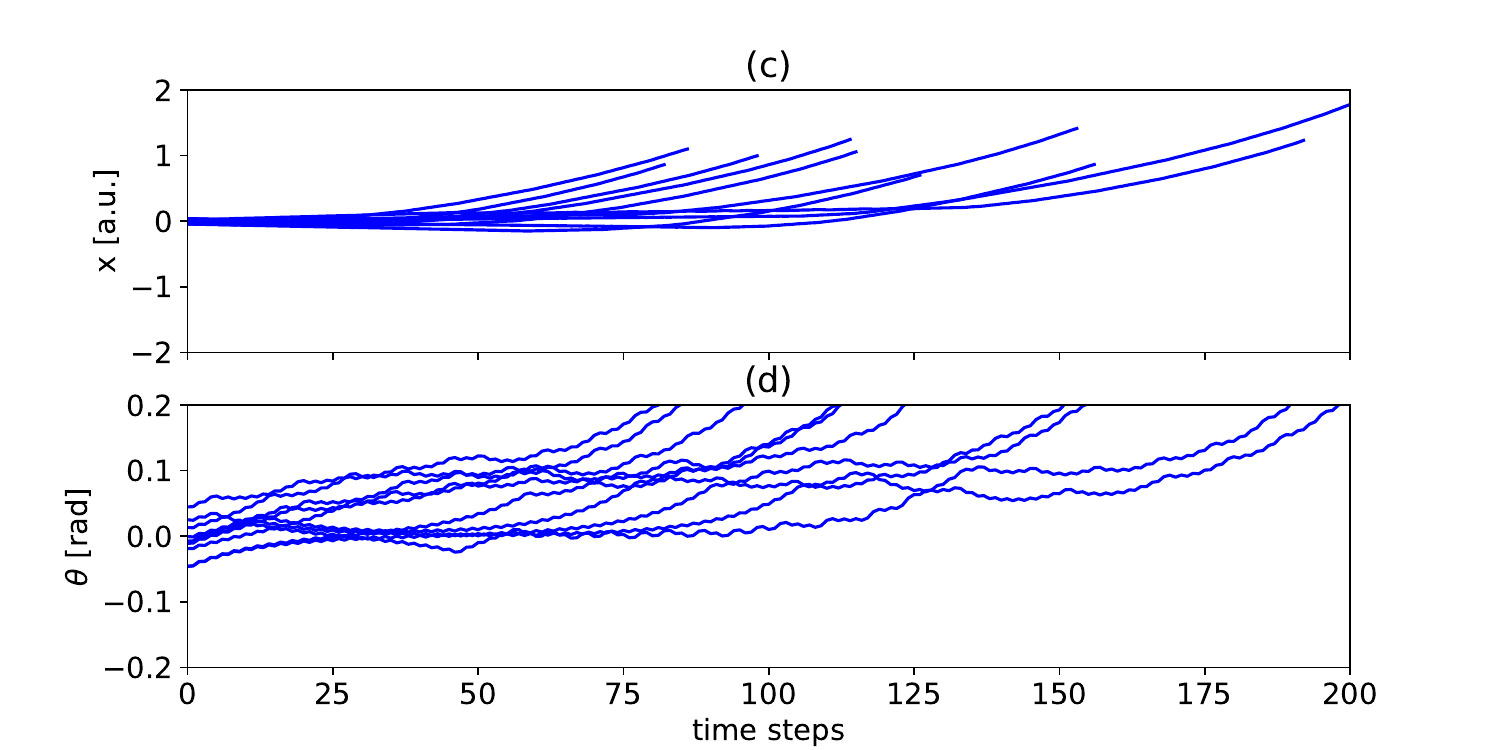}	
	\end{center}
	\caption{
		Time evolution of $x$ {\bf(a, c)} and $\theta$ {\bf(b, d)} under the balancing task, two specific controlled-tilt tasks, and a tilted-balancing task. {\bf(a, b)} Five exemplary traces per task. blue: balancing task; red: controlled tilt to the right with $\dot \theta^g = 0.5$; green: fast controlled tilt to the left with $\dot \theta^g = -5$. {\bf(c, d)}  10 exemplary traces for the tilted-balancing task with $\theta^g = 0.15$ rad.   
	}
	\label{fig:imit}
\end{figure}

One-shot imitation refers to the ability of a system "to learn from very few demonstrations of any given task, and instantly generalize to new situations of the same task, without requiring task-specific engineering" \cite{duan2017}. Here we formulate an extrinsic goal by presenting to the system a single sequence of target states, which are imprinted into its map. Imprinting means that the corresponding winning units are memorized as part of the goal. For example, if the goal is to balance the pole, a sequence of states $\V{u}_t = (x_t, \dot x_t, 0, 0), t = 0, 1, ..., T$ with upright stationary pole under various cart positions and velocities is presented to SapSom. Technically, instead of the true sequence of states, we only present the vector of expected values, $\V{u}^g = E_t[\V{u}_t]$, and the vector of inverse variances or \emph{precisions}, $\pi^g$, to the system (the superscript "g" stands for "goal"). 
For the pole balancing example, the set of goal values for $x$ and $\dot x$ will both show a large variance with zero mean. In contrast the goal values of $\theta$ and $\dot \theta$ are zero, resulting in zero mean and zero variance for these two variables. As the precisison is defined as the inverse variance, in this example the precisions of $x$ and $\dot x$ will be very small, the precisions of $\theta$ and $\dot \theta$ will be very high. We approximate the low precision values by zero and cap very high precision values to a maximum of 1, resulting in a mean vector of $\V{u}^g=(0, 0, 0, 0)$ and a precision vector of $\pi^g \approx (0, 0, 1, 1)$ for this example.

Reaching the target state then means to search for a sequence of actions, which drive the environment's actual state towards the target state(s) and keep it there. For simplicity, we avoid explicitly determining the distance between the actual state and all target states, but instead use the precision weighted distance between $\V{u}$ and $\V{u}^g$: $d(\V{u},\V{u}^g) = \sum_i \pi^g_i \; (u_i - u^g_i)^2 $. For action planning, one-step greedy forward search is applied i.e.  $\tau = 1$.

In this section we consider three different types of tasks, namely (i) the (pole) balancing task, (ii) the controlled-tilt task, and (iii) the tilted-balancing task, which are explained in the following. The balancing task consists of keeping the pole upright as stationary and as long as possible. As pointed out above, presenting the goal of a balanced pole results in imprinting $\V{u}^g=(0, 0, 0, 0), \; \pi^g = (0, 0, 1, 1)$. The controlled-tilt task consists of deliberately letting the pole tilt to a specified side, until the maximum tilt of $\theta^g = \pm0.2$ rad tolerated by the gym environment is reached. The challenge to the agent is to control the tilt process such that a specific angular velocity $\dot \theta^g$ is achieved at the instant where the border is hit. This task is assigned to the agent by imprinting $\V{u}^g=(0, 0, \theta^g, \dot \theta^g), \; \pi^g = (0, 0, 1, 1)$. In the tilted-balancing task, the goal is to keep the pole stationary, but in a tilted position $\theta^g \not=0$. This task, which is also not easy to achieve by humans, involves letting the pole tilt a little bit at the beginning, follwed by a controlled acceleration of the cart to the corresponding side in order to keep the tilt stationary afterwards.  This task is assigned to the agent by imprinting $\V{u}^g=(0, 0, \theta^g, 0), \; \pi^g = (0, 0, 1, 1)$. 

Fig \ref{fig:imit} shows the time evolution of cartpole location $x$ (Figs \ref{fig:imit}a and \ref{fig:imit}c) and angle $\theta$ (Figs \ref{fig:imit}b and \ref{fig:imit}d) for four specific goals given to the agent. The blue traces in Figs \ref{fig:imit}a and \ref{fig:imit}b show 5 attempts of the agent to solve the balancing task. The red traces were produced under a controlled tilt task with moderate goal angular velocity, $\V{u}^g=(0, 0, 0.2, 0.5)$. The green traces arose under another controlled-tilt task with high negative goal angular velocity $\V{u}^g=(0, 0, -0.2, -5)$.

The final angular velocities were $(0.57, 0.35, 0.40, 0.32, 0.33)$  for the moderate tilt case, and $(-2.57, -2.53, -2.46, -2.41, -2.58)$ for the rapid tilt case. The latter angular velocities are about the maximum which can be achieved when pushing to one side all the time, which was correctly predicted by the agent: All but one action over the five rapid tilt trials were "push right".   

Figs \ref{fig:imit}c and \ref{fig:imit}d illustrate how the system acts in a tilted-balancing task with goal angle $\theta^g = 0.15$ rad. To solve this task, the agent had to constantly accelerate the cart into the direction of the tilt in a controlled way, such that the cart leaves the regime previously explored and learned. The traces indicate that the agent manages to keep the pole steadily tilted over a considerable number of time steps, although it does not quite reach the goal of $\theta^g = 0.15$ rad but instead stabilizes values around $\theta = 0.08$ rad. Also, when approaching the screen boundaries, the agent fails to stabilize any longer, because this configuration is far from what it experienced during playful exploration with random actions only. 

From visual inspection of these examples, we find that the system is quite flexible in solving different related tasks and generalizes satisfactorily to previously unseen regimes. In the following, we quantitatively analyze the performance and variability thereof for a larger number of specific goals for these tasks.

\subsubsection{Performance analysis for the balancing task}

OpenAI defines "solving" the cartpole problem as being able to balance the pole over an average of 195 time steps per episode, taken over the last 100 episodes, where each episode ends in case of failure or after 200 time steps otherwise. In order to characterize the performance of the agent, we analyzed 100 episodes under the balancing task (Fig \ref{fig:imit}b, blue). We found that 61 episodes reached the limit of 200 steps of balancing, the average number of time steps was 188 steps. Hence, SapSom does not quite reach the definition for solving the task, but comes quite close. 

\vspace*{5mm}

\subsubsection{Performance analysis for the controlled-tilt task}

\begin{figure}[h!]
	\begin{center}
		
		\includegraphics[width=7cm]{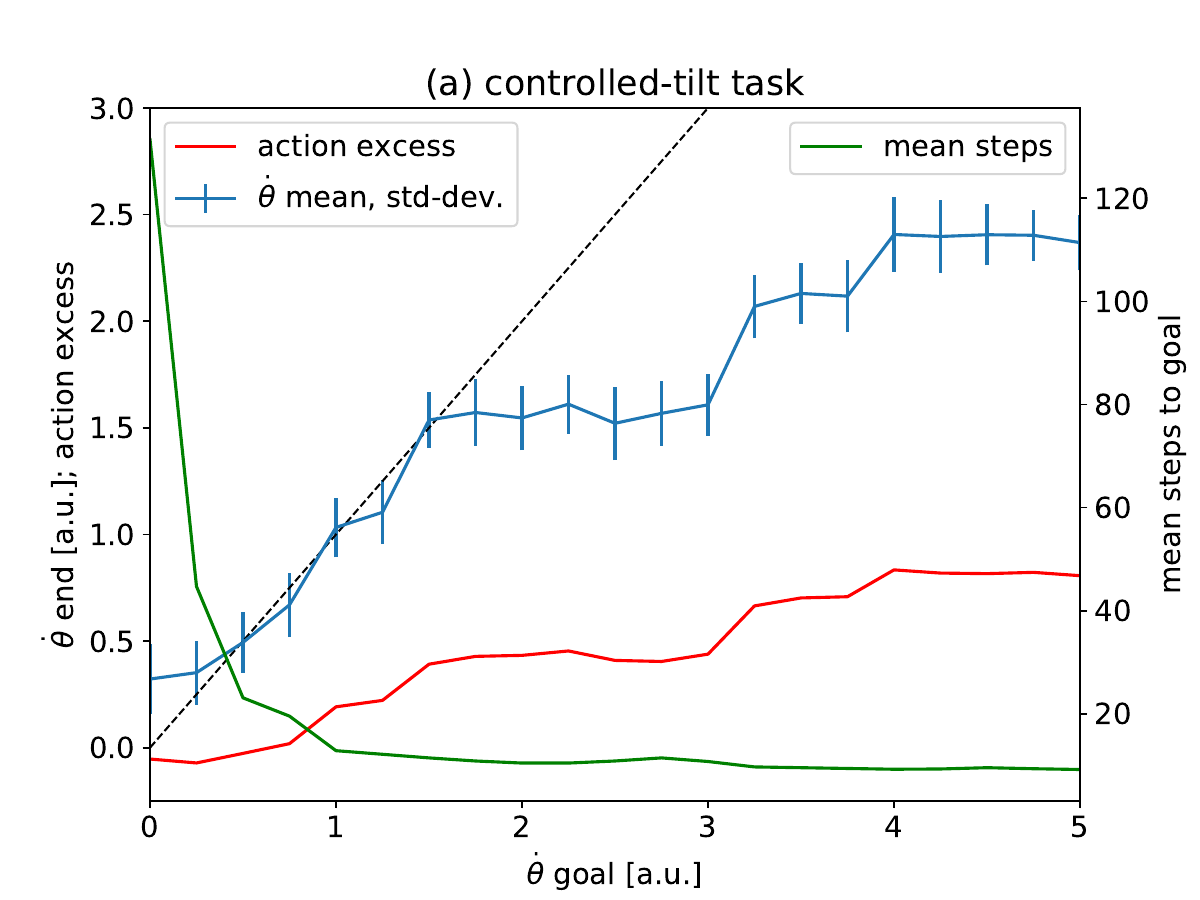}
		\includegraphics[width=7cm]{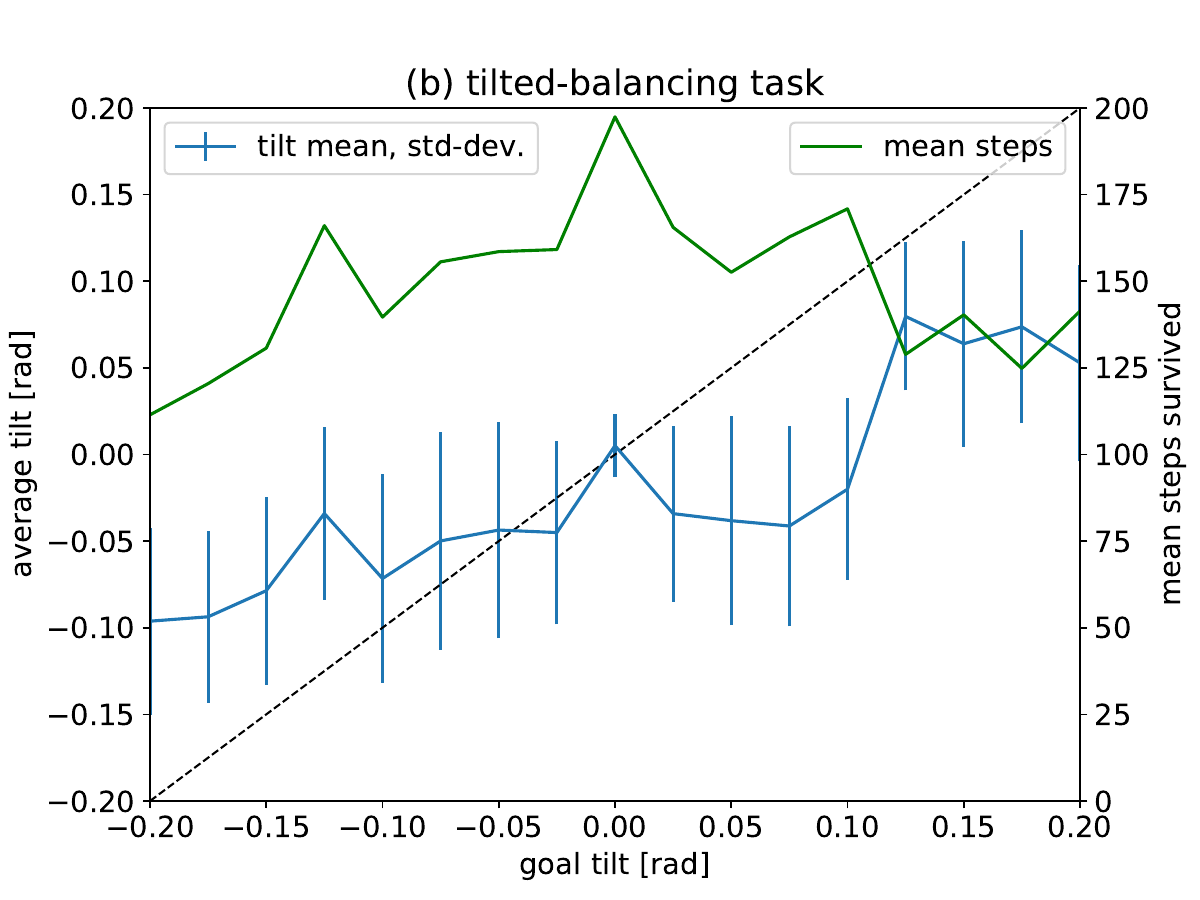}
		
	\end{center}
	\caption{
		{\bf (a)} Performance analysis for the controlled-tilt task. Blue trace: Means and standard deviations (vertical lines) over 20 runs of the actual final angular velocity as function of the goal angular velocity. Black dashed: bisector line. Red (left axis): Mean action excess. Green (right axis) Mean number of time steps until the done-signal. {\bf (b)} Performance analysis for the tilted-balancing task. Blue trace: Means and standard deviations (vertical lines) over 20 runs of the actual average tilt as function of the goal tilt. The actual average tilt was calcualted as the mean angle over time steps 50-100. Green (right axis) Mean number of steps until the done-signal 
	}
	\label{fig:perf}
\end{figure}

While keeping the goal angle constant at $\theta^g = 0.2$ rad, we systematically increased the goal angular velocity $\dot \theta$ from 0 to 5 in steps of 0.25. For each goal, 20 simulations with randomly initialized cartpoles were run and the mean and standard deviation of the actual angular velocities at the border (i.e. immediately after the done-signal of the cartpole environment) were computed. First of all we observed that in every trial the correct angle was approximated i.e. the agent always managed to tilt the pole to the right. The blue trace in Fig \ref{fig:perf}a plots the mean actual angular velocities and their standard deviations (vertical lines). In the range between 0.25 and 1.5 the goal angular velocity could be approximated very closely, whereas the performance dropped for higher goal values. To further analyze this situation, we plotted also the mean excess of the number of left push actions vs. right push actions, i.e. 
$(n_{left} - n_{right})/(n_{left} + n_{right})$ (Fig \ref{fig:perf}a, red line), and the mean number of time steps until the done signal was thrown (Fig \ref{fig:perf}a, green line, right axis). It can be seen that already at goal values of 1 and higher, the number of time steps until the border is reached is very small, namely about 10, and therefore there is only a limited number of possible counts of left vs right actions. In fact the steps in the actual final angular velocities correspond to steps in the action excess roughly at 0.4 (7 left vs. 3 right pushes), 0.6 (8 left vs. 2 right pushes), and 0.8 (9 left vs. 1 right push). Finally, the goal of zero angular velocity is slightly exceeded, however from the large number of steps survived it can be seen that the agent attempts to slowly approximate the border, and the negative action excess reflects its attempt to decelerate the tilt.

\subsubsection{Performance analysis for the tilted-balancing task}

While setting the goal angular velocity to zero, we systematically varied the goal angle from -0.2 to 0.2 in steps of 0.025 and ran the task 20 times for each goal. For each run, we computed the average tilt of the pole over time steps 50-100. This gave the agent some time to build up the tilt initially and excluded the later phase where many trials failed (cf also Fig \ref{fig:imit}c, d). Where the run survived less than 100 steps, the average was taken between 50 and the end of the run. The blue trace in Figure \ref{fig:perf}b plots the means and standard deviations of the average tilts over 20 runs as a function of the goal tilt (pole angle). While the absolute actual tilts systematically fall below the absolute goal tilts, there is a clear positive trend: the agent clearly attempts tilted balancing and on average succeeds in doing so. The larger standard deviations compared to Fig \ref{fig:perf}a reflect the higher difficulty of the tilted-balancing task as compared to controlled-tilt tasks. The green line (right axis) again displays the mean number of steps survived. For all, even the strong tilts, the agent manages to balance the pole for more than 100 steps on average. The highest survival time is achieved for zero tilt, which corresponds to the simple balancing task.

The results demonstrate, that SapSom can perform each of these tasks very well (although not perfectly) after only one presentation of the goal state. This is possible, because task solution is done via inference over the intuitive physics learned, rather than via slow modification of weights. Hence, in comparison with reinforcement learning, the presented system provides two major advantages: (i) it does not need any extrinsic reward structure (which has often to be engineered in a tedious process), and (ii) it can flexibly solve various tasks, which would require both a new reward structure and retraining in reinforcement learning. The latter flexibility has also been specified as an important feature of human-like learning and performance \cite{lake2017}.

\section{Discussion}
\label{sec:discussion}

The goal of this work was to suggest a minimal model which shows important properties of artificial intelligence: learning from experience, comprehension of its surroundings, reasoning, planning, and flexible solution of different tasks. It does so by  learning a representation of the dynamical physical properties of its environment by exploration, and by performing inference on this representation to reach goals. The philosophy behind this approach is related to the KISS principle and Occam's razor at the model level: we identified minimal ingredients 
which seem both plausible and important for achieving the mentioned properties (there may be completely different ways, though,  to generate the same behaviour). The key ingredients of SapSom are:
\begin{itemize}
	\item An adaptive and sparse sensorimotor representation, in particular the possibility to act on an environment rather than just observing, in order to learn by exploration.
	\item A temporal sequence learning and prediction mechanism at the sensorimotor level.
	\item A short term memory mechanism to store target states in state representation space.
	\item A mechanism to intrinsically generate action and state representations and to reason on their temporal evolution on the basis of temporal sequence prediction. 
	\item A distance measure between states. If reasoning is to be done at the representation level, a distance measure between sparse state representations, such as a topographic map, is required. 
\end{itemize} 

In the following, we first compare our model to a standard Q-learning setup for pole balancing, then discuss possible extensions of our model, and relate its mechanisms to biological findings.

\subsection{Comparison to state-of-the-art methods}
The proposed agent learns to achieve various different goals on the basis of a state-action and state-transition models which are learned in a completely unsupervised manner. Goals are imprinted by simple activation of states latent space, for example as a result of a single presentation of a target environmental state. Hence it is very easy to switch forth and back between different goals on a per-episode basis  for SapSom. In contrast, almost all state-of-the-art methods are related either to explicit physical modelling or to reinforcement learning (RL) in some sense (see Related Work section). While physical simulation engines require the human expert to design this engine, RL-methods, which might be most closely related to our approach,  heavily rely on adequate reward signals provided by the environment, which have to be manually designed by the human expert as well. Moreover, the characteristics of the reward signal provided by the environment usually encodes or encourages one specific goal only. A RL agent can only learn a new, different task, if (a) the experimenter handcrafts a new reward structure provided to the agent, and (b) the agent undergoes retraining in order to learn to exploit the new reward structure. For that reason, the high flexibility and generalization ability of SapSom alone renders our approach superior to physical engines and RL agents for the context addressed.

However, it is possible to compare the performance of SapSom to that of an RL-agent for a stationary task with defined reward structure. Therefore, we re-implemented standard tabular Q-learning \cite{sutton1998} for the cartpole environment following frequently used parameter settings. To approximate the number of $16 \times 16 = 256$ states of the SapSom agent, the continuous 4D state space $(x, \dot x, \theta, \dot \theta)$ was discretized to $3 \times 3 \times 6 \times 6 = 324$ bins for the Q-table, giving more emphasis to the important variables $\theta$ and $\dot \theta$. A learning rate of $\alpha = 0.1$ and an $\varepsilon$-greedy policy with $\varepsilon = 1-\log_{10}(n/25)$ constrained to $\varepsilon \in [0.1,1]$ for episode number $n\ge1$ was used.
As SapSom, the agent was trained over 4000 episodes. Of the last 1000 episodes, 87 percent reached the limit of 200 steps of balancing, the average number of time steps was 186. When comparing this to 61 percent completed episodes and 188 average steps of SapSom (see Method and Results section), we find that Sapsom reaches a performance comparable to Q-learning, but without manual binning of the input space, without taking any benefit of the reward signal, and while still being flexible enough to immediately perform a different task from the very next episode on. It can be concluded that our model is clearly more effective and efficient than Q-learning and probably related RL methods on the cartpole-task.


\subsection{Possible extensions}

A simple set of one-layer network architectures was used (in the full model, Fig \ref{fig:principle}a, a three layer architecture) to learn an embedding of the input space. Both the brain and state of the art representation techniques, in contrast, maintain hierarchical representations of modalities. In our model, single-layer state representation learning can easily be replaced by hierarchical self-organizing map structures \cite{miller2006}, or by contemporary high performance approaches such as (vector-quantized) variational autoencoders \cite{kingma2013, razavi2019}, generative adversarial networks, \cite{goodfellow2014} or deep convolutional architectures in general (e.g. \cite{krizhevsky2012}), which might be trained on raw frame sequences. A sparse representation at the embedding level, which is considered crucial for temporal sequence learning, can be either directly provided by such systems \cite{razavi2019}, or can be achieved by using a SOM or other competitive learning mechanism \cite{foeldiak1990} on top of their output or encoding layer.  

SapSom's temporal sequence learning mechanism is simply 1st order Markov, rendering it somewhat similar to Hidden Markov Models \cite{baum1966}. Clearly, reasoning on environments of natural complexity requires variable (and sometimes very high) order Markov representations. Hawkins et al. \cite{hawkins2009} developed a biologically plausible model for variable order sequence memory which is embedded in their hierarchical temporal memory framework. It is based on lateral cortical connections operating on a sparse representation, where each state is represented by multiple model neurons. Their approach could be most naturally incorporated in our framework, but also other approaches for variable order sequence prediction can be considered without changing the fundamental way the model works \cite{jensen2005,rawlinson2019}. A different obvious possibility to include sensitivity to variable length history is to use recurrent connectivity. Recurrent SOMs \cite{varsta1997, miller2006} can be trained to represent short sequences of input patterns instead of individual states.  

Besides being one-step forward only, action sequence planning is done simply by minimizing the precision-weighted euclidean distances between current and goal environmental states. For high dimensional state spaces, however, it is known that the euclidean distance can lose much of its meaning \cite{aggarwal2001}. Consequently, for high-dimensional spaces it might be beneficial to either resort to other distance measures, such as for example the cosine distance, or to put more weight on distance measurement in the topographically ordered latent space.

Another issue that is unresolved is how to stably establish a hierarchy of temporal timescales which is clearly present in human reasoning. When related to biology, assuming a "temporal timestep" within the brain of a few tens of milliseconds, all reasoning modeled here would operate in the sub second range. Humans, in contrast, make hierarchical action plans over a broad range from below seconds to years, and seem to compose longer term plans by abstracting from shorter ones \cite{khemlani2013}. Hierarchical arrangements of RSOMs have the potential to model a temporal hierarchy,  however, for longer history the algorithm requires some parameter finetuning and becomes subject to numerical instability. Temporal hierarchy might instead be more stably generated when interlacing RSOM-layers with a novelty-driven gating mechanism (Stetter, unpublished results).

In its minimal version, SapSom operates in completely unsupervised mode. It does not evaluate any supervisory or extrinsic reward signals. The feedback used is sensory information evoked by its own actions. Because there needs to be some intrinsic motivation mechanism which drives the system to explore its environment, to try and reach goals, and eventually to intrisically set goals, the present approach shows similarities with curiousity driven learning, where  intrinsic reward signals are generated based on prediction errors \cite{pathak2017}. Humans, in contrast, do use both extrinsic and intrinsically generated reward signals (e.g. \cite{schultz1997}). 

Reinforcement learning mechanisms can be incorporated in a natural way on top of the sensorimotor representation, as SapSom - up to the reward signal - considers its environment as a Markov decision process which it learns to approximate. 
The present approach therefore shows a natural link with model-based RL \cite{botvinick2014}.
Alternatively, Q-learning could use its update rule to optimize a Q-value that is assigned to each $sa$ unit in the sensorimotor map (Fig \ref{fig:principle}). The resulting policy would then replace the greedy forward search algorithm established here. Generally, however, finding efficient search strategies in action sequence space that approach human performance in being creative and solving problems is an important ongoing research issue.

\subsection{Relation to biology}

Conceptually, our model is placed in the middle of the spectrum ranging from approaches that deliberately abstract from the way the brain works \cite{russell2009} to computational neuroscience models which put their main focus on how cognitive mechanisms are specifically implementated in biological neural networks
\cite{deco2005}. Our approach is to formulate computational models, which are inspired by fundamental mechanisms of brain function, and are in principle compliant with what is known about biological information processing. Accordingly, a few parallels can be drawn between SapSom's present implementation and the brain's biological neural networks. 

First of all, Kohonen's self organizing maps are biologically motivated by retinotopy and smooth feature maps found in the early visual pathway, by mexican-hat like lateral cortical connectivity, and Hebbian learning. These aspects seem more closely related to biology than an error backpropagation mechanism, for which no biological correlate could be found so far. Moreover, SOMs have been used very successfully as models of self-organization in the early visual pathway \cite{obermayer1990}. Given the observation that all areas in the neocortex are laid out very uniformly and seem to perform similar operations \cite{mountcastle1978}, hierarchical systems of SOMs appear to be good candidates of neurally inspired models at least of posterior neocortex.

The state transition matrices optimized during prediction learning can be interpreted as lateral (Fig \ref{fig:principle}b) or top-down (Fig \ref{fig:principle}a) cortico-cortical connections. When doing so, the learning rule eq. (\ref{eq:predlearn}) just corresponds to spike-time dependent plasticity (e.g. \cite{bi1998}). 
For this we recall that that map state probabilities $p(\V{s})$ are derived from normalized activations of state neurons, eq. (\ref{eq:prob}).  When $T_a(\V{s'}, \V{s})$ is interpreted as connection from state-action neuron with index $\V{s},\; a$ to target neuron $\V{s'}$, the learning rule reads $\Delta T_a(\V{s'}, \; \V{s}) = (p_{t+1}(\V{s'}) - \hat p_{t+1}(\V{s'})) \cdot p_t(\V{s},a)$.  the first term says that $T_a$ will be increased, when $\V{s'}$ becomes active after $\V{s},a$ (stronger than expected), and is decreased, if $\V{s'}$ is expected active but remains silent. 

Moreover, there is evidence that prefrontal cortex is capable of actively maintaining information by robust, persistent neural activity, which, according to prominent models, might be rapidly updatable by a striatum-driven gating mechanism \cite{oreilly2006}. The system comprised of dorsolateral, anterior cingulate and orbitofrontal cortices, which interact with the striatum and thalamus, are considered crucial for representing goals and context, generating action plans and evaluating expected rewards thereof (for a review of data and detailled computational models, see \cite{oreilly2019}).  
Hence, non-distractible sustained activation might be a neural correlate of the model's goal states, whereas their distractible counterparts might underly working memory required during mental execution of the search through action plans. 

Learning more about the principles of this orchestration process of cortical states, or, in SapSom terminology, learning about how search and prediction should be ideally designed given an implicit world model, might lead to a better  understanding of the principles of human thinking in general -- an exciting field for future research.  

\section{Conflicts of Interest}
The authors declare that there is no conflict of interest regarding the publication of this paper.

\section{Acknowledgments}
The authors wish to thank Monika Stetter for numerous valuable discussions on the subject. MS was on sabbatical leave joining the CIML Lab at University of Regensburg. 


%

\bibliography{litdb}{}

\begin{thebibliography}{10}

\bibitem{aggarwal2001}
Charu~C. Aggarwal, Alexander Hinneburg, and Daniel~A. Keim.
\newblock On the surprising behavior of distance metrics in high dimensional
  space.
\newblock In Jan Van~den Bussche and Victor Vianu, editors, {\em Database
  Theory --- ICDT 2001}, pages 420--434, Berlin, Heidelberg, 2001. Springer
  Berlin Heidelberg.

\bibitem{ajay2019}
Anurag Ajay, Maria Bauza, Jiajun Wu, Nima Fazeli, Joshua~B. Tenenbaum, Alberto
  Rodriguez, and Leslie~P. Kaelbling.
\newblock Combining physical simulators and object-based networks for control.
\newblock Technical Report arXiv:1904.06580 [cs.RO], 2019.

\bibitem{battaglia2016}
Peter Battaglia, Razvan Pascanu, Matthew Lai, Danilo Jimenez~Rezende, and koray
  Kavukcuoglu.
\newblock Interaction networks for learning about objects, relations and
  physics.
\newblock In D.~D. Lee, M.~Sugiyama, U.~V. Luxburg, I.~Guyon, and R.~Garnett,
  editors, {\em Advances in Neural Information Processing Systems 29}, pages
  4502--4510. Curran Associates, Inc., 2016.

\bibitem{battaglia2018}
Peter~W. Battaglia, Jessica~B. Hamrick, Victor Bapst, Alvaro Sanchez-Gonzalez,
  Vinicius Zambaldi, Mateusz Malinowski, Andrea Tacchetti, David Raposo, Adam
  Santoro, Ryan Faulkner, Caglar Gulcehre, Francis Song, Andrew Ballard, Justin
  Gilmer, George Dahl, Ashish Vaswani, Kelsey Allen, Charles Nash, Victoria
  Langston, Chris Dyer, Nicolas Heess, Daan Wierstra, Pushmeet Kohli, Matt
  Botvinick, Oriol Vinyals, Yujia Li, and Razvan Pascanu.
\newblock Relational inductive biases, deep learning, and graph networks.
\newblock Technical Report arXiv:1806.01261 [cs.LG], 2018.

\bibitem{battaglia2013}
Peter~W. Battaglia, Jessica~B. Hamrick, and Joshua~B. Tenenbaum.
\newblock Simulation as an engine of physical scene understanding.
\newblock {\em Proceedings of the National Academy of Sciences},
  110(45):18327--18332, 2013.

\bibitem{baum1966}
Leonard~E. Baum and Ted Petrie.
\newblock Statistical inference for probabilistic functions of finite state
  markov chains.
\newblock {\em Ann. Math. Statist.}, 37(6):1554--1563, 12 1966.

\bibitem{bi1998}
Guoqiang Bi and Mu-ming Poo.
\newblock Synaptic modifications in cultured hippocampal neurons: Dependence on
  spike timing, synaptic strength, and postsynaptic cell type.
\newblock {\em Journal of Neuroscience}, 18:10464--72, 01 1999.

\bibitem{botvinick2014}
Matthew~M Botvinick and Ari Weinstein.
\newblock Model-based hierarchical reinforcement learning and human action
  control.
\newblock {\em Philosophical Transactions of the Royal Society B: Biological
  Sciences}, 369, 2014.

\bibitem{chang2017}
Michael~B. Chang, Tomer Ullman, Antonio Torralba, and Joshua~B. Tenenbaum.
\newblock A compositional object-based approach to learning physical dynamics.
\newblock Technical Report arXiv:1612.00341 [cs.AI], 2017.

\bibitem{clark2016}
Andy Clark.
\newblock {\em Surfing Uncertainty: Prediction, Action, and the Embodied Mind}.
\newblock Oxford University Press, Oxford, 2016.

\bibitem{deco2005}
Gustavo Deco and Edmund Rolls.
\newblock Attention, short-term memory, and action selection: A unifying
  theory.
\newblock {\em Progress in neurobiology}, 76:236--56, 08 2005.

\bibitem{duan2017}
Yan Duan, Marcin Andrychowicz, Bradly Stadie, Jonathan Ho, Jonas Schneider,
  Ilya Sutskever, Pieter Abbeel, and Wojciech Zaremba.
\newblock One-shot imitation learning.
\newblock In I.~Guyon, U.~V. Luxburg, S.~Bengio, H.~Wallach, R.~Fergus,
  S.~Vishwanathan, and R.~Garnett, editors, {\em Advances in Neural Information
  Processing Systems 30}, pages 1087--1098. Curran Associates, Inc., 2017.

\bibitem{ehrhardt2019b}
Sebastien Ehrhardt, Aron Monszpart, Niloy Mitra, and Andrea Vedaldi.
\newblock Taking visual motion prediction to new heightfields.
\newblock {\em Computer Vision and Image Understanding}, 181:14 -- 25, 2019.

\bibitem{ehrhardt2019}
Sebastien Ehrhardt, Aron Monszpart, Niloy Mitra, and Andrea Vedaldi.
\newblock Unsupervised intuitive physics from visual observations.
\newblock Technical Report arXiv:1805.05086 [cs.CV], 2019.

\bibitem{ehrhardt2017}
Sebastien Ehrhardt, Aron Monszpart, Niloy~J. Mitra, and Andrea Vedaldi.
\newblock Learning a physical long-term predictor.
\newblock Technical Report arXiv:1703.00247 [cs.AI], 2017.

\bibitem{foeldiak1990}
Peter F\"{o}ldi\'{a}k.
\newblock Forming sparse representations by local anti-hebbian learning.
\newblock {\em Biol. Cybern.}, 64(2):165 -- 170, 1990.

\bibitem{fragkiadaki2017}
Katerina Fragkiadaki, Jonathan Huang, Alex Alemi, Sudheendra Vijayanarasimhan,
  Susanna Ricco, and Rahul Sukthankar.
\newblock Motion prediction under multimodality with conditional stochastic
  networks.
\newblock Technical Report arXiv:1705.02082 [cs.CV], 2017.

\bibitem{friston2003}
Karl~J. Friston.
\newblock Learning and inference in the brain.
\newblock {\em Neural Networks}, 16(9):1325 -- 1352, 2003.

\bibitem{friston2005}
Karl~J. Friston.
\newblock A theory of cortical responses.
\newblock {\em Philosophical Transactions of the Royal Society B: Biological
  Sciences}, 360:815 -- 836, 2005.

\bibitem{friston2010}
Karl~J. Friston.
\newblock The free-energy principle: a unified brain theory?
\newblock {\em Nature reviews. Neuroscience}, 11:127--138, 2010.

\bibitem{fuster1971}
Joaquin~M. Fuster and Garrett~E. Alexander.
\newblock Neuron activity related to short-term memory.
\newblock {\em Science}, 173(3997):652--654, 1971.

\bibitem{goodfellow2016}
Ian~J. Goodfellow, Yoshua Bengio, and Aaron Courville.
\newblock {\em Deep Learning}.
\newblock MIT Press, Cambridge, MA, USA, 2016.

\bibitem{goodfellow2014}
Ian~J. Goodfellow, Jean Pouget-Abadie, Mehdi Mirza, Bing Xu, David
  Warde-Farley, Sherjil Ozair, Aaron Courville, and Yoshua Bengio.
\newblock Generative adversarial nets.
\newblock In {\em Proceedings of the 27th International Conference on Neural
  Information Processing Systems - Volume 2}, NIPS14, pages 2672 -- 2680,
  Cambridge, MA, USA, 2014. MIT Press.

\bibitem{graves2004}
Alex Graves, Douglas Eck, Nicole Beringer, and Juergen Schmidhuber.
\newblock Biologically plausible speech recognition with {LSTM} neural nets.
\newblock In Auke~Jan Ijspeert, Masayuki Murata, and Naoki Wakamiya, editors,
  {\em Biologically Inspired Approaches to Advanced Information Technology},
  pages 127--136, Berlin, Heidelberg, 2004. Springer Berlin Heidelberg.

\bibitem{graves2013}
Alex Graves, Abdel-rahman Mohamed, and Geoffrey~E. Hinton.
\newblock Speech recognition with deep recurrent neural networks.
\newblock Technical Report arXiv:1303.5778 [cs.NE], 2013.

\bibitem{hawkins2009}
Jeff Hawkins, Dileep George, and Jamie Niemasik.
\newblock Sequence memory for prediction, inference and behaviour.
\newblock {\em Philosophical Transactions of the Royal Society B: Biological
  Sciences}, 364(1521):1203--1209, 2009.

\bibitem{hohwy2013}
Jakob Hohwy.
\newblock {\em The Predictive Mind}.
\newblock Oxford University Press, Oxford, 2013.

\bibitem{jensen2005}
Steven Jensen, Daniel Boley, Maria Gini, and Paul Schrater.
\newblock Rapid on-line temporal sequence prediction by an adaptive agent.
\newblock {\em Proceedings of the International Conference on Autonomous
  Agents}, pages 67--73, 2005.

\bibitem{kansky2017}
Ken Kansky, Tom Silver, David~A. Mély, Mohamed Eldawy, Miguel
  Lázaro-Gredilla, Xinghua Lou, Nimrod Dorfman, Szymon Sidor, Scott Phoenix,
  and Dileep George.
\newblock Schema networks: Zero-shot transfer with a generative causal model of
  intuitive physics.
\newblock Technical Report arXiv:1706.04317 [cs.AI], 2017.

\bibitem{khemlani2013}
Sangeet~Suresh Khemlani, Robert Mackiewicz, Monica Bucciarelli, and Philip~N.
  Johnson-Laird.
\newblock Kinematic mental simulations in abduction and deduction.
\newblock {\em Proceedings of the National Academy of Sciences},
  110(42):16766--16771, 2013.

\bibitem{kingma2013}
Diederik~P {Kingma} and Max {Welling}.
\newblock Auto-encoding variational bayes.
\newblock Technical Report arXiv:1312.6114 [stat.ML], 2013.

\bibitem{kohonen1982}
Teuvo Kohonen.
\newblock Self-organized formation of topologically correct feature maps.
\newblock {\em Biological Cybernetics}, 43(1):59--69, January 1982.

\bibitem{krizhevsky2012}
Alex Krizhevsky, Ilya Sutskever, and Geoffrey~E Hinton.
\newblock Imagenet: classification with deep convolutional neural networks.
\newblock In F.~Pereira, C.~J.~C. Burges, L.~Bottou, and K.~Q. Weinberger,
  editors, {\em Advances in Neural Information Processing Systems 25}, pages
  1097--1105. Curran Associates, Inc., 2012.

\bibitem{lake2015}
Brenden~M. Lake, Ruslan Salakhutdinov, and Joshua~B. Tenenbaum.
\newblock Human-level concept learning through probabilistic program induction.
\newblock {\em Science}, 350:1332--1338, 2015.

\bibitem{lake2017}
Brenden~M. Lake, Tomer~D. Ullman, Joshua~B. Tenenbaum, and Samuel~J. Gershman.
\newblock Building machines that learn and think like people.
\newblock {\em Behavioral and Brain Sciences}, 40:e253, 2017.

\bibitem{miller2006}
Jeffrey~W. {Miller} and Peter~H. {Lommel}.
\newblock {\em {Biomimetic sensory abstraction using hierarchical quilted
  self-organizing maps}}, volume 6384 of {\em Society of Photo-Optical
  Instrumentation Engineers (SPIE) Conference Series}, page 63840A.
\newblock 2006.

\bibitem{mnih2015}
Volodymyr Mnih, Koray Kavukcuoglu, David Silver, Andrei~A. Rusu, et~al.
\newblock Human-level control through deep reinforcement learning.
\newblock {\em Nature}, 518(7540):529--533, February 2015.

\bibitem{morasso1997}
Pietro Morasso, Vittorio Sanguinetti, and Gino Spada.
\newblock A computational theory of targeting movements based on force fields
  an topology representing networks.
\newblock {\em Neurocomputing}, 15:411--434, 1997.

\bibitem{mountcastle1978}
Vernon~B. Mountcastle.
\newblock An organizing principle for cerebral function: {T}he unit module and
  the distributed system.
\newblock In F.~O. Schmitt, editor, {\em Neuroscience, Fourth Study Program},
  pages 21--42. MIT Press, Cambridge, MA, 1979.

\bibitem{nguyen2020}
H.~{Nguyen}, J.~{Patravali}, F.~{Li}, and A.~{Fern}.
\newblock Learning intuitive physics by explaining surprise.
\newblock In {\em 2020 IEEE/CVF Conference on Computer Vision and Pattern
  Recognition Workshops (CVPRW)}, pages 1539--1542, 2020.

\bibitem{obermayer1990}
Klaus Obermayer, Helge Ritter, and Klaus Schulten.
\newblock A principle for the formation of the spatial structure of cortical
  feature maps.
\newblock {\em PNAS}, 87(21):8345--8349, 1990.

\bibitem{oreilly2006}
Randall~C O'Reilly.
\newblock Biologically based computational models of high-level cognition.
\newblock {\em Science}, 314(5796):91--94, October 2006.

\bibitem{oreilly2019}
Randall~C. O'Reilly, Jacob Russin, and Seth~A. Herd.
\newblock Computational models of motivated frontal function.
\newblock In Mark D'Esposito and Jordan~H. Grafman, editors, {\em The Frontal
  Lobes}, volume 163 of {\em Handbook of Clinical Neurology}, pages 317 -- 332.
  Elsevier, 2019.

\bibitem{pathak2017}
Deepak Pathak, Pulkit Agrawal, Alexei~A. Efros, and Trevor Darrell.
\newblock Curiosity-driven exploration by self-supervised prediction.
\newblock In {\em Proceedings of the 34th International Conference on Machine
  Learning - Volume 70}, ICML 2017, pages 2778--2787. JMLR.org, 2017.

\bibitem{pearl2000}
Judea Pearl.
\newblock {\em Causality: Models, Reasoning, and Inference}.
\newblock Cambridge University Press, Cambridge, 2000.

\bibitem{pearl2018}
Judea Pearl.
\newblock Theoretical impediments to machine learning with seven sparks from
  the causal revolution.
\newblock Technical Report arXiv:1801.04016 [cs.LG], 2018.

\bibitem{rao1998}
Rajesh Rao and Dana Ballard.
\newblock Predictive coding in the visual cortex: a functional interpretation
  of some extra-classical receptive-field effects.
\newblock {\em Nature Neuroscience}, 2:79--87, 02 1999.

\bibitem{rawlinson2019}
David Rawlinson, Abdelrahman Ahmed, and Gideon Kowadlo.
\newblock Learning distant cause and effect using only local and immediate
  credit assignment.
\newblock Technical Report arXiv:1905.11598 [stat.ML], 2019.

\bibitem{rawlinson2012}
David Rawlinson and Gideon Kowadlo.
\newblock Generating adaptive behaviour within a memory-prediction framework.
\newblock {\em PLoS ONE}, 7:e29264, 2012.

\bibitem{razavi2019}
Ali Razavi, Aaron van~den Oord, and Oriol Vinyals.
\newblock Generating diverse high-fidelity images with {VQ-VAE-2}.
\newblock Technical Report arXiv:1906.00446 [cs.LG], 2019.

\bibitem{riochet2018}
Ronan Riochet, Mario Castro, Mathieu Bernard, Adam Lerer, Rob Fergus,
  Véronique Izard, and Emmanuel Dupoux.
\newblock Intphys: A framework and benchmark for visual intuitive physics
  reasoning.
\newblock Technical Report arXiv:1803.07616 [cs.AI], 2018.

\bibitem{russell2009}
Stuart Russell and Peter Norvig.
\newblock {\em Artificial Intelligence: A Modern Approach}.
\newblock Prentice Hall Press, USA, 3rd edition, 2009.

\bibitem{schmidhuber2010}
Jurgen Schmidhuber.
\newblock Artificial scientists \& artists based on the formal theory of
  creativity.
\newblock {\em Artificial General Intelligence - Proceedings of the Third
  Conference on Artificial General Intelligence, AGI 2010}, 06 2010.

\bibitem{schultz1997}
Wolfram Schultz, Peter Dayan, and P~Read Montague.
\newblock A neural substrate of prediction and reward.
\newblock {\em Science}, 275(5306):1593--1599, 1997.

\bibitem{smith2019}
Kevin Smith, Lingjie Mei, Shunyu Yao, Jiajun Wu, Elizabeth Spelke, Josh
  Tenenbaum, and Tomer Ullman.
\newblock Modeling expectation violation in intuitive physics with coarse
  probabilistic object representations.
\newblock In H.~Wallach, H.~Larochelle, A.~Beygelzimer, F.~d'Alch\'{e} Buc,
  E.~Fox, and R.~Garnett, editors, {\em Advances in Neural Information
  Processing Systems 32}, pages 8985--8995. 2019.

\bibitem{spelke2007}
E.~S Spelke and K.~D. Kinzler.
\newblock Core knowledge.
\newblock {\em Developmental Science}, 10:89--96, 2007.

\bibitem{stetter2006}
Martin Stetter.
\newblock Dynamic functional tuning of nonlinear cortical networks.
\newblock {\em Phys. Rev. E}, 73:031903, Mar 2006.

\bibitem{stetter2021}
Martin Stetter and Elmar~W. Lang.
\newblock Learning intuitive physics and one-shot imitation using
  state-action-prediction self-organizing maps.
\newblock Technical Report arXiv:2007.01647 [cs.AI], 2021.

\bibitem{sutton1998}
Richard~S. Sutton and Andrew~G. Barto.
\newblock {\em Reinforcement Learning: An Introduction}.
\newblock The MIT Press, second edition, 2018.

\bibitem{toussaint2006}
M.~{Toussaint}.
\newblock A sensorimotor map: Modulating lateral interactions for anticipation
  and planning.
\newblock {\em Neural Computation}, 18(5):1132--1155, 2006.

\bibitem{vansteenkiste2018}
Sjoerd van Steenkiste, Michael Chang, Klaus Greff, and Jürgen Schmidhuber.
\newblock Relational neural expectation maximization: Unsupervised discovery of
  objects and their interactions.
\newblock Technical Report arXiv:1802.10353 [cs.LG], 2018.

\bibitem{varsta1997}
Markus Varsta, Jukka Heikkonen, and Jose del R.~Millan.
\newblock Context learning with the self-organizing map.
\newblock {\em Proceedings of the Workshop on Self-Organizing Maps '97}, pages
  197--202, 1997.

\bibitem{wang_j2018}
Jane Wang, Zeb Kurth-Nelson, Dharshan Kumaran, Dhruva Tirumala, Hubert Soyer,
  Joel Leibo, Demis Hassabis, and Matthew Botvinick.
\newblock Prefrontal cortex as a meta-reinforcement learning system.
\newblock {\em Nature Neuroscience}, 21:860--868, 06 2018.

\bibitem{wang_j2016}
Jane~X Wang, Zeb Kurth-Nelson, Dhruva Tirumala, Hubert Soyer, Joel~Z Leibo,
  Remi Munos, Charles Blundell, Dharshan Kumaran, and Matt Botvinick.
\newblock Learning to reinforcement learn.
\newblock Technical Report arXiv:1611.05763 [cs.LG], 2016.

\bibitem{wang_y2019}
Yaqing Wang, Quanming Yao, James Kwok, and Lionel~M. Ni.
\newblock Generalizing from a few examples: A survey on few-shot learning.
\newblock Technical Report arXiv:1904.05046 [cs.LG], 2019.

\bibitem{watters2017}
Nicholas Watters, Daniel Zoran, Theophane Weber, Peter Battaglia, Razvan
  Pascanu, and Andrea Tacchetti.
\newblock Visual interaction networks: Learning a physics simulator from video.
\newblock In I.~Guyon, U.~V. Luxburg, S.~Bengio, H.~Wallach, R.~Fergus,
  S.~Vishwanathan, and R.~Garnett, editors, {\em Advances in Neural Information
  Processing Systems 30}, pages 4539--4547. 2017.

\bibitem{wu2017}
Jiajun Wu, Erika Lu, Pushmeet Kohli, William~T Freeman, and Joshua~B Tenenbaum.
\newblock Learning to see physics via visual de-animation.
\newblock In {\em Advances in Neural Information Processing Systems 30}, pages
  153--164, 2017.

\bibitem{wu2015}
Jiajun Wu, Ilker Yildirim, Joseph~J Lim, Bill Freeman, and Josh Tenenbaum.
\newblock Galileo: Perceiving physical object properties by integrating a
  physics engine with deep learning.
\newblock In C.~Cortes, N.~D. Lawrence, D.~D. Lee, M.~Sugiyama, and R.~Garnett,
  editors, {\em Advances in Neural Information Processing Systems 28}, pages
  127--135. 2015.

\bibitem{wu_t2018}
Tailin Wu, John Peurifoy, Isaac~L. Chuang, and Max Tegmark.
\newblock Meta-learning autoencoders for few-shot prediction.
\newblock Technical Report arXiv:1807.09912 [cs.LG], 2018.

\bibitem{ye2018}
Tian Ye, Xiaolong Wang, James Davidson, and Abhinav Gupta.
\newblock Interpretable intuitive physics model.
\newblock In Vittorio Ferrari, Martial Hebert, Cristian Sminchisescu, and Yair
  Weiss, editors, {\em Computer Vision - {ECCV} 2018 - 15th European
  Conference, Munich, Germany, September 8-14, 2018, Proceedings, Part {XII}},
  volume 11216 of {\em Lecture Notes in Computer Science}, pages 89--105.
  Springer, 2018.

\bibitem{zheng2018}
David Zheng, Vinson Luo, Jiajun Wu, and Joshua~B. Tenenbaum.
\newblock Unsupervised learning of latent physical properties using
  perception-prediction networks.
\newblock Technical Report arXiv:1807.09244 [cs.LG], 2018.

\end{thebibliography}
\bibliographystyle{plain}

\end{document}